\documentclass[acmsmall,nonacm]{acmart}

\setcopyright{none}  
\renewcommand\footnotetextcopyrightpermission[1]{} 
\usepackage[T1]{fontenc}
\usepackage[utf8]{inputenc}
\usepackage{times}

\usepackage{microtype} 

\usepackage{graphicx}
\usepackage{float}
\usepackage{subcaption}
\usepackage{threeparttable}
\usepackage{longtable}
\usepackage{lscape}

\usepackage{xcolor}
\usepackage{pifont}

\usepackage{tikz}
\usetikzlibrary{shapes.geometric, arrows.meta, positioning, trees, shadows}
\usepackage{forest}

\usepackage{amsmath}
\usepackage{booktabs}
\usepackage{array}
\usepackage{multirow}
\usepackage{multicol}

\usepackage{hyperref}
\usepackage{fontawesome5}

\acmMonth{8}

\begin{document}

\title{The Rise of Small Language Models in Healthcare: A Comprehensive Survey}

\author{Muskan Garg}
\email{garg.muskan@mayo.edu}
\affiliation{
  \institution{Artificial Intelligence \& Informatics, Mayo Clinic}
  \city{Rochester}
  \state{Minnesota}
  \country{USA}
}

\author{Shaina Raza}
\email{shaina.raza@torontomu.ca}
\affiliation{
 \institution{Vector Institute}
 \city{Toronto}
 \country{Canada}}

 \author{Shebuti Rayana}
\email{rayanas@oldwestbury.edu}
\affiliation{
  \institution{SUNY at Old Westbury}
  \city{Old Westbury, New York}
  \country{USA}
}

\author{Xingyi Liu}
\email{liu.xingyi@mayo.edu}
\affiliation{
  \institution{Artificial Intelligence \& Informatics, Mayo Clinic}
  \city{Rochester}
  \state{Minnesota}
  \country{USA}
}

\author{Sunghwan Sohn}
\email{sohn.sunghwan@mayo.edu}
\affiliation{%
  \institution{Artificial Intelligence \& Informatics, Mayo Clinic}
  \city{Rochester}
  \state{Minnesota}
  \country{USA}
}

\renewcommand{\shortauthors}{Garg et. al.,}

\begin{abstract}
Despite substantial progress in healthcare applications driven by large language models (LLMs), growing concerns around data privacy, and limited resources; the small language models (SLMs) offer a scalable and clinically viable solution for efficient performance in resource-constrained environments for next-generation healthcare informatics. Our comprehensive survey presents a taxonomic framework to identify and categorize them for healthcare professionals and informaticians. The timeline of healthcare SLM contributions establishes a foundational framework for analyzing models across three dimensions: NLP tasks, stakeholder roles, and the continuum of care. We present a taxonomic framework to identify the architectural foundations for building models from scratch; adapting SLMs to clinical precision through prompting, instruction fine-tuning, and reasoning; and accessibility and sustainability through compression techniques. Our primary objective is to offer a comprehensive survey for healthcare professionals, introducing recent innovations in model optimization and equipping them with curated resources to support future research and development in the field. Aiming to showcase the groundbreaking advancements in SLMs for healthcare, we present a comprehensive compilation of experimental results across widely studied NLP tasks in healthcare to highlight the transformative potential of SLMs in healthcare. The updated repository is available at Github\footnote{\href{https://github.com/drmuskangarg/SLMs-in-healthcare/}{https://github.com/drmuskangarg/SLMs-in-healthcare/}}.
\end{abstract}


\maketitle

\section{Background}
\label{sec:1}


The global small language model (SLM) market is projected to reach USD 29.64 billion by 2032, with the healthcare industry anticipated to experience the fastest compound annual growth rate (CAGR) of 18.31\% from 2024 to 2032\footnote{\href{https://www.globenewswire.com/news-release/2025/03/12/3041492/0/en/Small-Language-Model-Market-to-Reach-USD-29-64-Billion-by-2032-SNS-Insider.html}{https://www.globenewswire.com/}}. This growth is driven by the growing adoption of SLM in medical diagnosis, patient care, and administrative processes. The use of SLM in analyzing extensive medical literature and patient data provides clinicians with evidence-based interpretations to support clinical decision support system (CDSS)\footnote{\href{https://www.healthcare.digital/single-post/small-language-models-slms-have-the-potential-to-revolutionise-healthcare-ai-in-addition-to-llms}{https://www.healthcare.digital/}}. 

A rapid evolution of transformer-based architectures has catalyzed a paradigm shift in natural language processing (NLP), with large (or very large) language models (LLMs) exceeding hundreds of billions of parameters (e.g., Med-PaLM~\cite{tu2024towards}), achieving state-of-the-art performance on various benchmarks \cite{singhal2025toward}. However, the computational and energy demands associated with the training, fine-tuning and deployment of LLMs in production require GPU clusters and significant power consumption at the megawatt scale, driving substantial interest in SLMs in both research and healthcare applications \cite{popov2025overview}. Healthcare is a broad, overarching system that encompasses the prevention, diagnosis, treatment, and management of illness, along with the preservation of physical and mental well-being through services offered by medical, nursing, and allied health professionals.

\subsection{Motivation}
\label{motivation}
The deployment of language models in hospital settings demands an uncertain trade-off among high-stakes accuracy, regulatory compliance, and computational efficiency \cite{singh2023centering}. While LLMs effectively capture clinical concepts \cite{fu2023clinical}, their high \textit{inference latency} (>500ms per query) and \textit{memory demands} (>16GB VRAM) make them unsuitable for real-time analysis of medical records with protected health information (PHI). 
Healthcare data is governed by stringent regulations such as \textit{HIPAA and GDPR} and requires on-premise processing to mitigate privacy risks \cite{jonnagaddala2025privacy}. \textit{Regulatory agencies} (e.g., FDA\footnote{\href{https://www.fda.gov/}{https://www.fda.gov/}}) encourages transparent AI-driven decision-making to promote interpretation. Knowledge-driven models generate interpretable results based on trusted external medical knowledge bases, supporting regulatory compliance such as \textit{EU Medical Device Regulation (MDR)} \cite{gilbert2023european}. Training LLMs cause an immense carbon footprint of more than 500 metric tons $\text{CO}_2$ which is not \textit{sustainable} for environmental goals. In contrast, SLM is optimized for \textit{watts per inference}, a sustainability metric \cite{potkins2024improve}. They reduce energy consumption by 10 to 100 times through optimization and compression techniques, addressing the \textit{Jevons paradox in green AI} (making models more energy-efficient can sometimes lead to an overall increase in resource consumption—not a decrease, as one might expect) \cite{luccioni2025efficiency}.
Critical care scenarios require \textit{subsecond decision making} driven by clinical NLP (cNLP) tasks where LLMs with high inference latencies are not suitable for \textit{time-sensitive applications} \cite{bai2024beyond}. As such, we opt to expose the potential of SLM to address specialized healthcare challenges under limited resources and time constraints. 

\subsection{Language model fundamentals}
SLM models lie on the "knee" of \textit{Pareto frontier}, a curve that represents the set of solutions to multi-objective optimization where no objective can be improved without worsening at least one other objective \cite{cho2025cosmosfl}, maximizing the utility per parameter. These models with fewer parameters than LLMs, aim for \textit{Pareto efficiency}, a change that makes at least one person better off without making anyone else worse off. This balances performance with minimal use of resources, reducing latency, memory consumption and environmental impact. Consequently, SLMs are optimized for production-ready deployment, which is achieved through architectural innovations and task-aware specializations \cite{van2024survey}. Unlike LLMs, SLMs in healthcare are designed as \textit{democratization} (Accessible, participatory, and equitable model development) tools \cite{subramonian2024understanding}, enabling decision-making in low-resource settings and shaping a design philosophy that extends beyond model size. Platforms such as Pytorch\footnote{\href{https://pytorch.org/serve/llm_deployment.html}{https://pytorch.org/}}, Hugging Face\footnote{\href{https://huggingface.co/}{https://huggingface.co/}} and Lamini\footnote{\href{https://www.lamini.ai/}{https://www.lamini.ai/}} define the practical deployability of language models that run on consumer hardware without specialized infrastructure.

\textbf{Transformers overview}: SLMs are transformer-based models that capture contextual relationships and dependencies in language. A transformer is a deep learning model that processes sequences in parallel by converting tokenized words into numerical embeddings. Transformers use \textit{positional encoding} to maintain word order and allow parallel processing. 
The transformer encoder converts input sequences into context-rich representations, while the decoder generates output sequences. Transformers are categorized into encoder-only, decoder-only, and encoder-decoder architectures. The encoder captures relationships among words using self-attention mechanisms in the attention layer, assigning weights based on contextual relevance \cite{belhaouari2025efficient,liu2023self}. Encoder-only models (BERT-like) are generally good in discriminative tasks. 
Decoder-only models (GPT-like) generate text by predicting the next token in a sequence, making them ideal for generative tasks. Encoder-decoder models (e.g., BART \cite{lewis2019bart}) combine input understanding and output generation. Given our focus on generative tasks in hospital settings, we emphasize decoder-only SLMs for their efficiency and adaptability. 

\subsection{Scope}
\label{scope}

Early studies define SLM as the models with 1B to 20B parameters, while LLMs have hundreds of billion parameters \cite{subramanian2025small}. Industry research defines SLMs based on deployment limits, such as models running on CPU-only edge devices~\cite{sun2020mobilebert}. Later research suggests a relative definition, where SLMs are 10 to 100 times smaller than the largest models of their time \cite{jovanovic2024compacting}. The definition of ``\textit{small}” in SLMs is relative in time and evolves with hardware advancements. It generally refers to reduced computational complexity (e.g., parameter count) and lower storage requirements (e.g., model size). 

\textbf{Parameter range:} SLMs typically fall within a spectrum in which the lower bound represents the minimum size required to exhibit emergent abilities for healthcare applications, while the upper bound is constrained by the maximum size manageable under limited resource conditions \cite{lu2024small}. While models exceeding 7B parameters typically require high-end GPUs, 7B represents a practical upper limit for local execution; our study defines SLM to be up to 7B parameter range, though future models may surpass this.

\textbf{Search strategy}: The existing literature used mixed terminology for referring to the SLM: local language models, remote language models, compressed models, optimized models, on-device language models, small LLMs and lightweight language models. We used these keywords to identify SLMs designed for NLP-driven CDSS from well-known sources: PubMed, ACM, IEEE, Google Scholar, Scopus. Our study covers three dimensions: (i) NLP tasks, (ii) stakeholders, (iii) care continuum. The NLP tasks covers question answering (QA), named entity recognition (NER), relation extraction (RE), information extraction (IE), natural language inference/ understanding (NLI/ NLU), text classification (CLS), summarization (SUM). We focus on three important stakeholders: patient, healthcare professionals, research/material. The care continuum covers triaging, diagnosis, treatment, recovery, risk prediction, and virtual health assistants. 

\begin{figure}[!ht]
    \centering
    \begin{subfigure}[b]{0.49\textwidth}
        \centering
        \includegraphics[width=\textwidth]{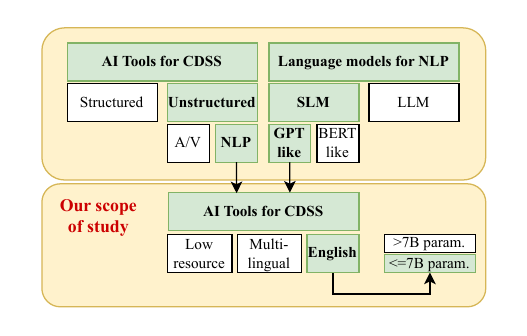}
        \caption{Scope of our study}
        \label{figscopea}
    \end{subfigure}
    \hfill
    \begin{subfigure}[b]{0.50\textwidth}
        \centering
        \includegraphics[width=\textwidth]{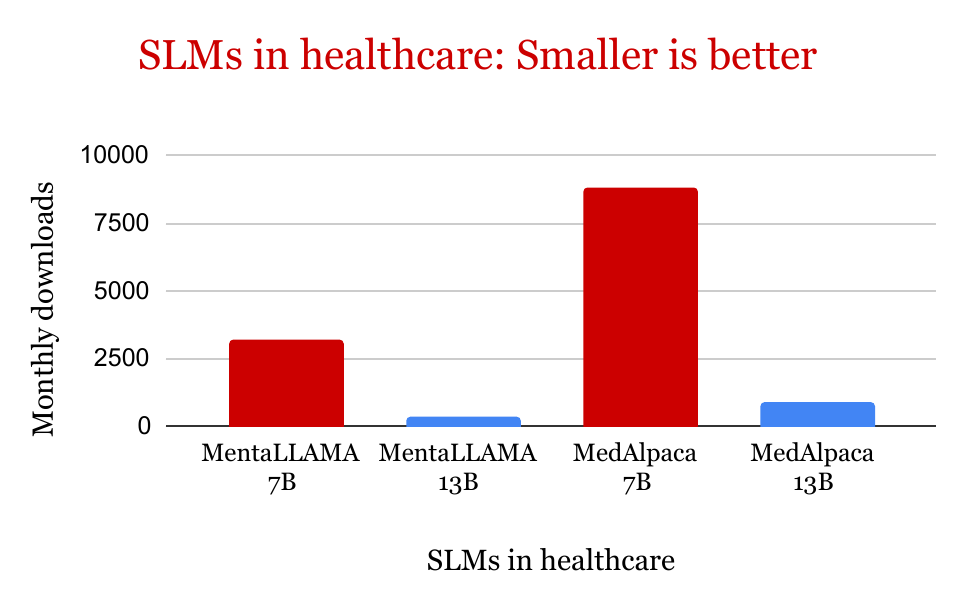}
        \caption{Monthly downloads}
        \label{figdownload}
    \end{subfigure}
    \caption[Study rationale]{Opening remarks of the study rationale: (a) Defined scope of the study to investigate the NLP-centered SLM development for healthcare professionals; (b) Monthly downloads for 7B and 13B parameters model from Huggingface for both MentaLLAMA and MedAlpaca.}
    \label{fig:scope}
\end{figure}

\textbf{Inclusion criteria.} Unlike 
\textit{fundamental} bidirectional BERT-like transformers, our interest lies in the decoder-only transformer architecture models (traced back to GPT) in the healthcare domain with open weights. We include SLMs trained on English language data for cNLP tasks, developed before April 2025. The scope of our study is defined in Fig. \ref{figscopea}. \textbf{Exclusion criteria.} We exclude models that are non-English, designed for inferring non-textual audiovisual data, multimodal SLMs, agentic AI systems, and models exceeding 7 billion parameters. Additionally, models lacking peer-reviewed publications for SLM models are not considered in this study.

\begin{figure}
\scriptsize
\centering

\tikzset{
    basic/.style  = {draw, text width=.4cm, align=center, font=\sffamily, rectangle},
    root/.style   = {basic, rounded corners=2pt, thin, align=center, fill=green!30},
    onode/.style = {basic, thin, rounded corners=2pt, align=center, fill=green!60, text width=2cm},
    tnode/.style = {basic, thin, align=left, fill=pink!60, text width=6.5cm},
    unode/.style = {basic, thin, align=left, fill=pink!60, text width=8cm},
    xnode/.style = {basic, thin, rounded corners=2pt, align=center, fill=blue!10, text width=2cm},
    ynode/.style = {basic, thin, rounded corners=2pt, align=center, fill=yellow!20, text width=2cm},
    wnode/.style = {basic, thin, align=left, fill=pink!10!blue!80!red!10, text width=2cm},
    edge from parent/.style={draw=black, edge from parent fork right}
}

\resizebox{\textwidth}{!}{
\begin{forest} for tree={
    grow=east,
    growth parent anchor=west,
    parent anchor=east,
    child anchor=west,
    rounded corners, 
    edge path={\noexpand\path[\forestoption{edge},->, >={latex}] 
         (!u.parent anchor) -- +(0pt,0pt) |-  (.child anchor) 
         \forestoption{edge label};}
}
[{\rotatebox{90}{SLMs in Healthcare}}, basic, l sep=5mm, anchor=center,
    [SLM in healthcare: Outlook \S\ref{outlook}, xnode, l sep=5mm,
        [Assessment methods \S\ref{eval}, ynode, l sep=5mm,  
            [\textbf{Safety} \cite{bhimani2025real,elangovan2024lightweight,liu2025lookahead,han2024medsafetybench}\textbf{, Trustworthiness} \cite{labrak2024biomistral,xie2023faithful,aljohani2025comprehensive}\textbf{, Memory and energy efficiency} \cite{husom2024price}, tnode]]
        [The role of SLM in healthcare \S\ref{role}, ynode, l sep=5mm,
            [\textbf{Data curation} \cite{shi2024mentalqlm}\textbf{, On-device personalization} \cite{qin2024enabling}\textbf{, Ensembling} \cite{cho2025cosmosfl,gondara2025elm}\textbf{, distillation} \cite{shi2024mentalqlm}, tnode]]
        [Current state of the art \S\ref{current}, ynode, l sep=5mm,
            [ \textbf{MedQA} \cite{bolton2024biomedlm,labrak2024biomistral,fu2024biomistral,chen2023meditron,wang2024apollo,wu2304pmc}\textbf{, Language understanding} \cite{gutierrez2023biomedical,labrak2024biomistral,chen2023meditron}\textbf{, Mental health} \cite{shi2024mentalqlm,yang2024mentallama,kim2024mhgpt}, tnode]]]
    [Compression techniques: Ensuring accessibility and sustainability \S\ref{compression}, xnode, l sep=5mm,
        [Quantization \S\ref{quantization}, ynode, l sep=5mm,  
            [ \textbf{PTQ}
            \cite{kumar2024mental,shi2024mentalqlm}
            \textbf{, QAT}
            \cite{bolton2024biomedlm,kim2024mhgpt}
            \textbf{, AWQ}
            \cite{labrak2024biomistral}, tnode]]   
        [Pruning \S\ref{pruning}, ynode, l sep=5mm,  
            [ \textbf{Structured pruning} \cite{lu2024all, ma2023llm}
            \textbf{, Unstructured pruning} \cite{zhang2024pruning}, tnode]]
        [Knowledge distillation \S\ref{knowdistil}, ynode, l sep=5mm,  
            [ \textbf{Black box KD} \cite{zheng2024sleepcot}\textbf{, White box KD} \cite{ding2024distilling,liu2024large,pavel2024non,zhang2024llm,vedula2024distilling}\textbf{, Adapt and Distill} \cite{yao2021adapt}, tnode]]]
    [Adapting SLMs to clinical precision \S\ref{icttuning}, xnode, l sep=5mm,
        [Supervised fine tuning \S\ref{finetuning}, ynode, l sep=5mm,  
            [ \textbf{Instruction Tuning} \cite{zhang2023instruction,nazar2025design,tran2024bioinstruct,wu2024instruction}\textbf{, Instruction Fine Tuning} \cite{zhang2023alpacare,li-etal-2024-llamacare,luo2024taiyi,liu2024evaluating,ranjit2024rad,namvar5098278predicting,leong2024efficient,rollman2025practical}\textbf{, PEFT} \cite{shi2024mentalqlm,goswami2024parameter,zhou2023llm,abishek2024bio,wang2023medical}, tnode]]    
        [In context learning \S\ref{ict}, ynode, l sep=5mm,  
            [\textbf{CoT} \cite{wang2024atscot,jiang2025comt}\textbf{, RAG and external knowledge} \cite{wu2025integrating,li2024rt,min2021metaicl}\textbf{, Metacognition} \cite{church2023does,esteitieh2025towards,griot2025large}, tnode]
            [ \textbf{Preferential alignment} \cite{zhang2025iimedgpt,gururajan2024aloe,liubest}\textbf{, Prompting mechanisms} \cite{ahmed2024med,diao2023active,wu2025automedprompt}, tnode]]]
    [Architectural foundations: Building SLMs from scratch \S\ref{arch}, xnode, l sep=5mm,
        [Attention mechanism \S\ref{attention}, ynode, l sep=5mm,  
            [ \textbf{Multi head attention} \cite{luo2022biogpt}\textbf{, Flash attention} \cite{bolton2024biomedlm, labrak2024biomistral}, tnode]] 
        [Pretraining components \S\ref{pretrain}, ynode, l sep=5mm,  
            [ \textbf{Custom tokenization} \cite{bolton2024biomedlm,kim2024mhgpt}\textbf{, Clinical abbreviation disambiguation} \cite{hosseini2024leveraging}\textbf{, Vocabulary scaling} \cite{huang2025over,tao2024scaling,yu2025scaling}, tnode]]
        [Healthcare data \S\ref{healthcaredata}, ynode, l sep=5mm,  
            [\textbf{Data quality} \cite{wang2024apollo,sallinen2025llama,shi2024mentalqlm}\textbf{, Importance resampling} \cite{xie2023data}\textbf{, Data diversity} \cite{shi2024mentalqlm}, tnode]]
        [SLM in healthcare: Timeline and configuration \S\ref{config}, ynode, l sep=5mm,  
            [\textbf{NLP for medical conditions} \cite{li2024cancergpt,li2024cancerllm,ranjit2024rad,ranjit2024radphi}\textbf{, Healthcare NLP} \cite{lu2025med,fu2024biomistral,labrak2024biomistral,han2023medalpaca,wang2024apollo,yang2024clinicalmamba,wu2304pmc,zhang2023alpacare,chen2023meditron,luo2022biogpt}\textbf{, Mental health and conversations} \cite{shi2024mentalqlm,kim2024mhgpt,yang2024mentallama,liu2023chatcounselor,li2023chatdoctor}, tnode]]]
]
\end{forest}
}
\caption[Taxonomy]{Taxonomic overview of the study design for the rise of SLMs in healthcare.}
\label{tree}
\end{figure}
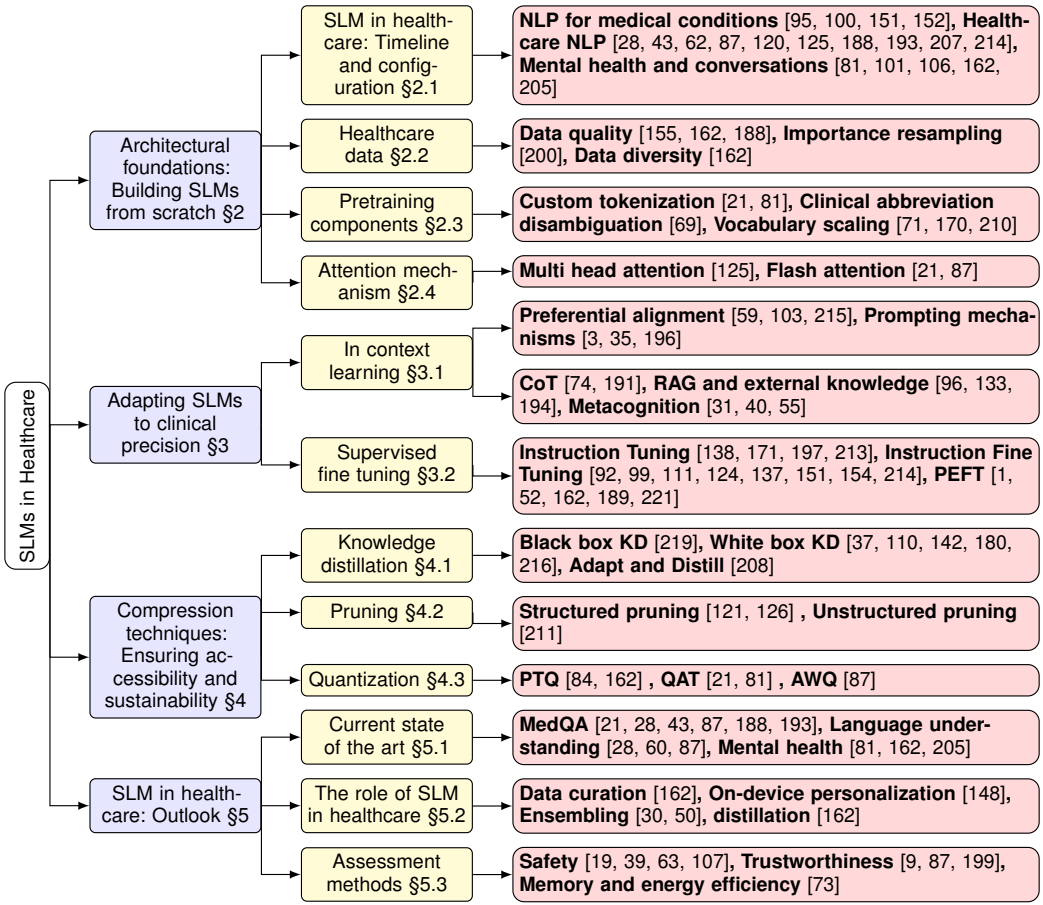

\subsection{Contributions}
\label{contributions}
The rise of healthcare SLM is evident from Fig. \ref{figdownload}. Existing reviews and surveys cover general studies on LLM \cite{wu2025survey}, the application of LLM in healthcare \cite{guo2024large,wang2024large}, and broad analyses of SLM \cite{lu2024small,van2024survey,wang2024comprehensive}. Despite the abundance of existing surveys and reviews, as shown in Table \ref{tabsurvey}, there remains a pressing need for a comprehensive examination of the rise of small language models (SLMs) for generative AI applications in healthcare tasks. To address this gap, our survey provides a comprehensive analysis of SLMs, highlighting their optimization strategies, clinical applicability, and transformative potential in resource-limited and safety-critical healthcare settings. To the best of our knowledge, this is the first comprehensive survey on SLMs in healthcare.  Our major contributions are as follows: 
\begin{enumerate}
    \item Propose an intuitive taxonomy for tuning and compression strategies in language models, and review prior studies to highlight emerging trends in model development (Fig. \ref{tree}).
    \item Outline the timeline of the SLM development in healthcare to categorize them by parameter range and functionality (Fig. \ref{figtimeline}).
    \item Compile existing open-source resources, including publicly available models (Table \ref{tabl:comparison}) and commonly used training datasets (Table \ref{tabdatasets}) in the literature.
    \item Examine reductions in carbon footprint and showcase healthcare-specific SLMs to guide researchers in adopting sustainable and efficient practices in the healthcare research domain (Table \ref{tab:carbon}).
    \item Outline the role of SLMs in healthcare applications (\S\ref{role}), examine associated challenges, and highlight key research directions for healthcare integration (\S\ref{discussion}).
\end{enumerate}

We first identified the models through our search strategy, discovered the model configurations, and other architectural foundations. We further investigate its adaptation to clinical precision and compression techniques, followed by outlook as shown in Fig. \ref{tree}.

\begin{table}
\scriptsize
\centering
\begin{tabular}{p{7.5cm}|>{\centering\arraybackslash}m{01 cm}|>{\centering\arraybackslash}m{0.5cm}>{\centering\arraybackslash}m{0.5cm}>{\centering\arraybackslash}m{0.5cm}>
{\centering\arraybackslash}m{0.5cm}>{\centering\arraybackslash}m{0.5cm}}
\toprule
\textbf{Existing studies}  & \textbf{Year} & \multicolumn{5}{c}{\textbf{Focus}} \\ 
& &  \textbf{HS} 
 & \textbf{NLP} 
 & \textbf{SLM}  & \textbf{MD} & \textbf{D/G}\\ 
\hline
\midrule
Small language models: survey, measurements, and insights \cite{lu2024small} & 2024 & \ding{55} & \ding{51} & \ding{51}  & \ding{55} & \ding{51} \\
A survey of small language models \cite{van2024survey} & 2024 & \ding{55} & \ding{51} & \ding{51}  & \ding{55} & \ding{51} \\
A comprehensive survey of small language models in the era of large... \cite{wang2024comprehensive}  & 2024 & \ding{55} & \ding{51} & \ding{51}  & \ding{55} & \ding{51} \\
A survey of personalized large language models...
\cite{liu2025survey}& 2025 & \ding{55} & \ding{51} & \ding{51}  & \ding{55} & \ding{51} \\
    A survey on LLM-generated text detection \cite{wu2025survey} & 2023 & \ding{55} & \ding{51} & \ding{55}  & \ding{55} & \ding{51}\\
    Localizing in-domain adaptation of transformer-based... \cite{buonocore2023localizing} & 2023 & \ding{51} & \ding{51} & \ding{51}  & \ding{55} & \ding{55}\\
    Efficient compressing and tuning methods for large language models... \cite{kim2024efficient} & 2023 & \ding{55} & \ding{51} & \ding{51}  & \ding{55} & \ding{51}\\ 
    Large language models in medical and... \cite{wang2024large} & 2024 & \ding{51} & \ding{51} & \ding{55}  & \ding{55} & \ding{55}\\ 
    Large language model for mental health... \cite{guo2024large}& 2024 & \ding{55} & \ding{51} & \ding{55}  & \ding{51} & \ding{51}\\
\midrule
\textsc{Our study}  & 2025 &  \ding{51} & \ding{51} & \ding{51} & \ding{51} & \ding{51} \\
    \bottomrule
\end{tabular}
\caption{Focused existing surveys in this field. Here HS: Healthcare-specific, MD: Mental discourse, D/G: Decoder-only or GPT-like.}
\label{tabsurvey}
\end{table}

\section{Architectural foundations: Building SLMs from scratch}
\label{arch} 
The \textit{lightweight architectures} are designed for efficient performance with fewer parameters and reduced computational overhead while maintaining high performance in medical applications. While model efficiency is important, the success of SLMs in healthcare also depends on the quality and accessibility of medical data. This motivates a closer look at the data landscape in hospital settings. In this section, we identify SLMs in healthcare, their timeline, and configurations (\S\ref{config}) to focus on architectural foundations in healthcare through healthcare data (\S\ref{healthcaredata}), pre-training components (\S\ref{pretrain}), and attention mechanisms (\S\ref{attention}).

\subsection{SLM in healthcare: Timeline and configuration}
\label{config}

\begin{figure}[!ht]
    \centering
\includegraphics[width=0.97\textwidth]{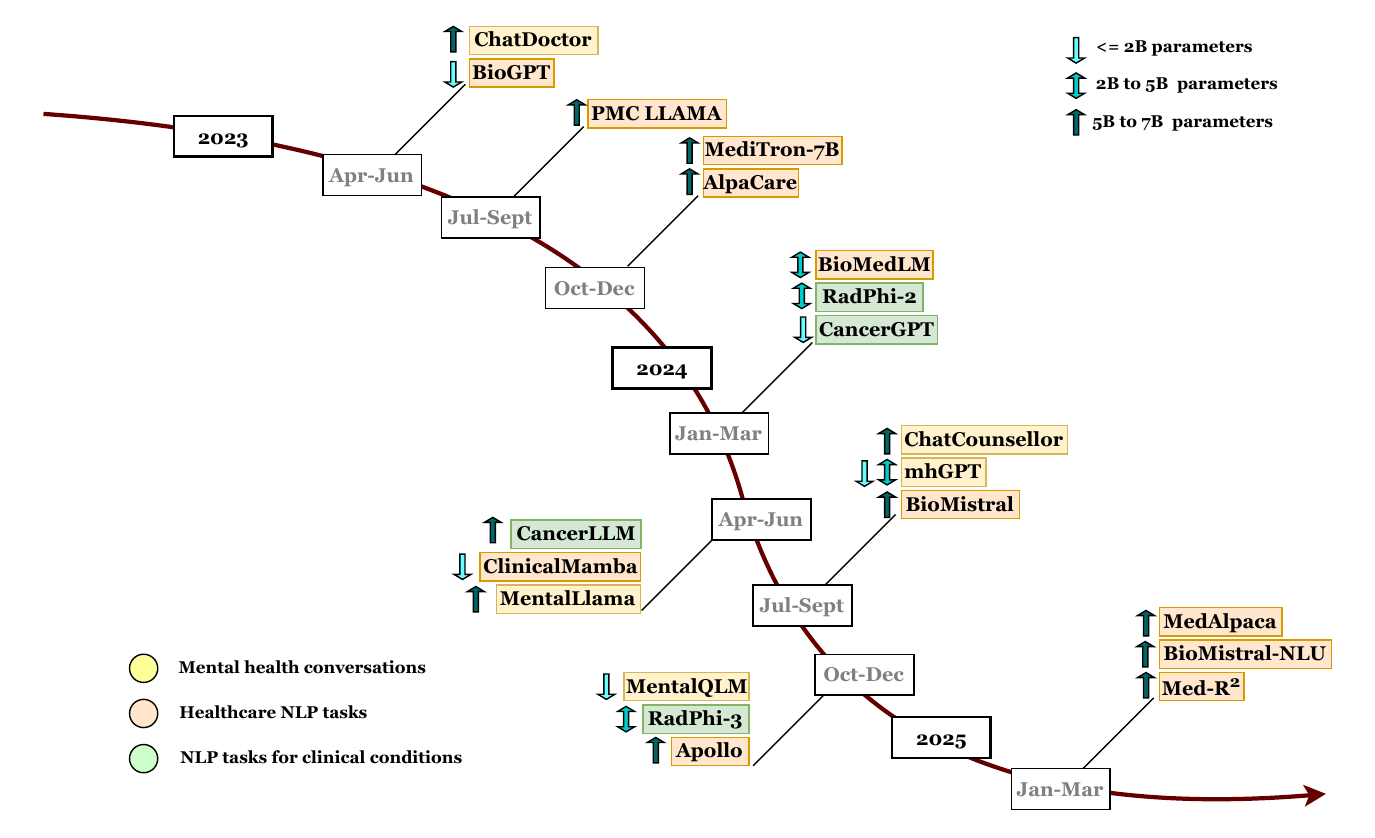}
    \caption{Timeline of SLM development in healthcare. Transforming NLP-driven clinical decision-making, NLP tasks for clinical conditions, and doctor-patient conversation or mental health analysis.}
    \label{figtimeline}
\end{figure}

We identified 20 SLMs in the literature based on our search criteria and selection principles. We present the timeline of model development in Fig. \ref{figtimeline}. SLMs are configured with different aspects to optimize their performance. Notably, very few (4 out of 20) studies provide detailed configurations of SLMs for healthcare applications. Table \ref{tab:slm_configs} presents the available configurations we identified, offering an overview of current development efforts. The model size, measured in parameters, typically ranges from 124 million to 7 billion weights. The model depth, defined by the number of transformer layers, typically spans from 22 to 32 layers. The hidden dimension, or the size of the model’s embedding vectors, usually ranges from 1024 to 4096 units. To optimize attention processing, SLMs employ 16 to 64 parallel attention heads, enabling the model to focus on various aspects of the data simultaneously. Additionally, the context window defines the maximum sequence length, allowing the model to process up to 4096 tokens, which is crucial for handling long clinical texts such as patient records and discharge summaries. These design choices ensure that SLMs achieve both high accuracy and efficiency in healthcare applications. 

\begin{table}
\scriptsize
\caption{Model configurations for existing SLM in healthcare, despite the limited availability in this domain. HS: hidden size, VS: Vocabulary size, MCW: Max context window}
\centering
\begin{tabular}{l|l|l|l|l|c|c|c|c|c}
\toprule
 \textbf{Model Name} & \textbf{Size} & \textbf{Year} & \textbf{Attention} & \textbf{Layers} & \textbf{HS} & \textbf{Heads} & \textbf{Activation} & \textbf{VS} & \textbf{MCW}\\
\toprule
BioGPT & 347M, 1.5B & 2023 & MHA & 24 & 1024 & 16 & GELU & 42384 & 1024\\
BioMedLM \cite{bolton2024biomedlm} & 2.7B & 2024 & FA & 32 & 2560 & 20 & GELU & 28896 & 1024  \\
BioMistral & 7B & 2024 & FA & 32 & 4096  & 32 & SiLU& 32000 & 4096 \\
\multirow{2}{*}{mhGPT} &  1.98B, & \multirow{2}{*}{2024} & \multirow{2}{*}{-} & \multirow{2}{*}{22} & \multirow{2}{*}{3072} & 64 & \multirow{2}{*}{-} & \multirow{2}{*}{52000} & \multirow{2}{*}{2048} \\
& 2.8B & & & & & 32 & & & \\ 
\bottomrule
\end{tabular}
\label{tab:slm_configs}
\end{table}

\subsection{Healthcare data}
\label{healthcaredata}
The healthcare domain presents unique challenges for language models due to the nature of its data. Clinical text data from EHRs, discharge summaries and physician notes holds valuable insights but is underutilized due to unstructured formats and medical jargon \cite{fu2023clinical}. While models like Med-PaLM \cite{tu2024towards} excel in medical NLP, their scale limits deployment, underscoring the need for efficient and reliable SLM training strategies. 

The datasets used for SLM construction are shown in Table \ref{tabdatasets}. \citet{yan2024large} highlights the challenges with data privacy and lack of the open domain datasets. A potential solution is the synthetic health data generation (SHDG), leveraging learning from statistical distributions of real-world datasets. Recent reviews and surveys on SHDG \cite{gonzales2023synthetic,pezoulas2024synthetic} provide a quick reference guide for selecting appropriate synthetic data generation strategies. Recent studies used LLMs for SHDG and validated that LLM-generated synthetic data can improve system performance \cite{kang2024synthetic}. A comprehensive study on metrics for SHDG evaluation \cite{osorio2024privacy} noted the absence of a composite metric for quality assessment. SHDG supports \textit{data enhancement}\footnote{augmenting, interpolating, or fusing with synthetic data} and \textit{de-identification}\footnote{masking personally identifiable information from patient records} in healthcare but also raises concerns about vulnerabilities such as \textit{data poisoning attacks}\footnote{malicious data is inserted into the training set, intentionally corrupting healthcare models to produce inaccurate or biased outputs} that can compromise model integrity. \textit{Data poisoning attack} \cite{zhao2025data} occurs when even 0.001\% of training tokens are replaced by medical misinformation \cite{alber2025medical}. 
\citet{lobo2022data} emphasized the need for robust evaluation against data poisoning, which can be mitigated through secure data practices, thorough cleaning, and multistage model training \cite{he2025multi}.

\begin{table}
\centering
\scriptsize
\begin{tabular}{@{} p{2.5cm} p{4cm} p{6.5cm}@{}}
\toprule
 Model & Training Dataset & Evaluation Dataset \\
\midrule
AlpaCare~\cite{zhang2023alpacare} & MedInstruct-52k & MedInstruct-test, iCliniq, MedQA, HeadQA, PubmedQA, MedMCQA, MeQSum, MMLU, BBH, TruthfulQA \\
Apollo \cite{wang2024apollo} & ApolloCorpora (12 multilingual med sources) & XMedBench (6 languages) \\
BioGPT~\cite{luo2022biogpt} & PubMed abstracts & BC5CDR, KD-DTI, DDI, PubMedQA, BioASQ, HoC \\
BioMedLM~\cite{bolton2024biomedlm} & PubMed abstracts and full articles & MedMCQA, MedQA, MMLU Medical Genetics, PubMedQA, BioASQ \\
BioMistral~\cite{labrak2024biomistral} & PubMed Central (3B tokens) & MMLU-medical, MedQA, MedMCQA, PubMedQA \\
BioMistral-NLU~\cite{fu2024biomistral} & MNLU-Instruct (33 datasets) & 15 NLU datasets (BLURB, BLUE) \\
CancerGPT~\cite{li2024cancergpt} & DrugComb (common tissue) & DrugComb(Drug pairs in rare tissue) \\
CancerLLM~\cite{li2024cancerllm} & Clinical notes, CancerNER, ICDdiagnosis & Phenotype extraction and diagnosis generation; counterfactual and misspelling robustness testbeds \\
ChatCounsellor \cite{liu2023chatcounselor} & Psych8k & Synthetic data from interviews \\
ChatDoctor~\cite{li2023chatdoctor} & Stanford Alpaca + HealthCareMagic & iCliniq, Wikipedia, MedlinePlus \\
ClinicalMamba~\cite{yang2024clinicalmamba} & MIMIC-III (ICU notes) & n2c2 2018, ICD-9 coding \\
Med-$R^2$~\cite{lu2025med} & None (prompt-based only) & MedQA, MedMCQA, MMLU-Med \\
MedAlpaca~\cite{han2023medalpaca} & Medical Meadow (Anki Flashcards, StackExchange, WikiDoc, CORD-19, PubmedQA, MedQA, OpenAssistant) & USMLE \\
Meditron~\cite{chen2023meditron} & GAP-Replay corpus & MedQA, PubMedQA, MMLU-Medical \\
MentaLlama~\cite{yang2024mentallama} & IMHI dataset (Reddit, Twitter, SMS) & Dreaddit, SAD, CAMS, Loneliness, MultiWD, IRF and 4 more \\
MentalQLM~\cite{shi2024mentalqlm} & DR, MultiWD, Dreddit, Irf, SAD & Held-out subsets of DR, MultiWD, Dreddit, Irf, SAD \\
mhGPT~\cite{kim2024mhgpt} & PubMed + Reddit (mental health) & IRF, Dreaddit, SAD, MultiWD, PPD-NER \\
PMC-LLAMA~\cite{wu2304pmc} & MedC-K (papers, textbooks) + MedC-I (instructions) & MedQA, MedMCQA, PubMedQA \\
RadPhi-2~\cite{ranjit2024rad} & Radiopaedia.org, Super-Natural Instructions, Mimic-CXR report data & Held-out Radiopaedia QA sets, MIMIC-CXR report data \\
RadPhi-3~\cite{ranjit2024radphi} & Mimic-CXR, MedicalDiff VQA, ChestImagenome, CheXpert Plus reports & RaLEs, RadGraph2, MEDNLI, RADNLI, PadChest \\

\bottomrule
\end{tabular}
\caption{Comparison of healthcare SLMs: Year and training/evaluation datasets.}
\label{tabslmdatasets}
\end{table}

\begin{table}
\scriptsize
\centering
\begin{tabular}{p{2.8cm}p{10cm}}
\toprule
\textbf{Task} & \textbf{Datasets} \\
\midrule
Language Inference & iCliniq, ICD9 Coding~\cite{mullenbach2018explainable}, i2b2~\cite{uzuner20112010}, MedlinePlus~\cite{li2023chatdoctor}, MIMIC-CXR~\cite{johnson2019mimic} \\
Language Understanding & Clinical Knowledge, College Biology, Medical Genetics, Professional Medicine~\cite{hendrycks2020measuring} \\
Named Entity Recognition & BC5CDR~\cite{li2016biocreative}, ICD9 coding~\cite{mullenbach2018explainable}, i2b2~\cite{uzuner20112010} \\
Question Answering & BioASQ~\cite{tsatsaronis2015overview}, MedMCQA~\cite{pal2022medmcqa}, MedQA-USMLE~\cite{kung2023performance}, PubMedQA~\cite{jin2019pubmedqa}, Radiopaedia QA~\cite{ranjit2024rad} \\
Relation Extraction & BC5CDR~\cite{li2016biocreative}, DDI~\cite{herrero2013ddi}, KD-DTI~\cite{hou2022discovering} \\
Text Classification & CAMS~\cite{garg2022cams}, Dreaddit~\cite{turcan2019dreaddit}, HoC~\cite{baker2016automatic}, IRF~\cite{garg2023annotated}, Loneliness~\cite{tsakalidis2019can}, MultiWD~\cite{sathvik2023multiwd}, SAD~\cite{mauriello2021sad} \\

\bottomrule
\end{tabular}
\caption{Open-source datasets used as benchmarks for NLP-driven healthcare tasks in the development of SLMs for clinical applications.}
\label{tabdatasets}
\end{table}

\textbf{High-quality datasets}: The \textit{`entropy law'} highlights that the compression ratio of the training data is a decisive factor affecting model performance, if the overall quality and consistency of selected samples remain unchanged \cite{yin2024entropy}. With limited computational resources, the use of high-quality data points is more effective for training than large and noisy datasets, improving efficiency with limited computational resources \cite{kang2024synthetic}. The data cleaning process filters noise, removes duplicates, handles outliers, and balances datasets to ensure better data quality. The discovery of data selection mechanisms ensures the learning of language models from high-quality data, thereby improving efficiency \cite{zhao2024clues}. However, poor selection may lead to biased results and reduced model performance \cite{xie2023data}. A recent survey on data selection mechanism highlights the need to scale down datasets, standardize evaluation metrics and data-centric benchmarking \cite{albalak2024survey}. \citet{qin2024unleashing} discussed data assessment and selection methods to identify the most beneficial data points that are different from standard data quality assessment measures \cite{lee2018framework}. 
 
\textbf{Data cleaning mechanisms} are de-identification, de-sensitization, and de-duplication, that enhance model performance by reducing redundancy and increasing diversity. \textit{De-duplication} improves data quality and prevents data leakage between training and test sets, as demonstrated by models like MediTron \cite{sallinen2025llama} and Apollo \cite{wang2024apollo}. \textit{Importance resampling} efficiently selects relevant data in a reduced feature space, outperforming existing methods and improving downstream model accuracy \cite{xie2023data}. Notably, \citet{shi2024mentalqlm} employed perplexity and k-center greedy methods to increase data quality and data diversity, respectively, while developing SLM. After surveying data diversity metrics, \citet{yang2025measuring} proposed an effective data selection strategy; however, its application in the healthcare domain remains unexplored, underscoring the need to benchmark selection methods for medical datasets to enhance diversity, minimize redundancy, and prevent overfitting. We enlist the training and evaluation datasets used in the existing SLMs for healthcare in Table \ref{tabslmdatasets} and mention open source datasets for each of the NLP task in healthcare (see Table \ref{tabdatasets}).
 
\subsection{Pretraining components}
\label{pretrain}
To effectively leverage SLMs, it is essential to understand the underlying components for pre-training the model.  
Pretraining on diverse, unlabeled corpora enables language models to learn general linguistic patterns, while task-specific fine-tuning allows them to adapt effectively to downstream NLP tasks in healthcare. Following data quality assessment and data selection, SLMs internalize linguistic structures, semantic relationships, and general world knowledge, establishing a robust foundation of language understanding.

\textbf{Tokenization}: Healthcare elements - ICD codes and drug names are challenging due to their ambiguity and domain-specific nature.
Word-level tokenization (e.g., Word2Vec) splits text into whole words, leading to a large vocabulary and poor handling of rare words. Character-level tokenization uses individual characters, which handles rare or misspelled words well but results in longer sequences and less linguistic structure. In contrast to both, \textit{subword tokenization} (e.g., BPE, WordPiece, and the Unigram Language Model) offers a middle ground by breaking words into frequent and meaningful subunits - prefixes or suffixes, mostly used by modern LLMs to efficiently handle both common and rare words \cite{bradshaw2025large}.
Tokenization of non-language text (like medical codes) prevents LLMs from recognizing the inherent structure of coding systems, making them less effective at accurate classification \cite{soroush2024large}. Recently, tokenization in clinical texts has improved by using domain-specific tools like abbreviation disambiguation \cite{hosseini2024leveraging} and medical knowledge bases (e.g., Unified Medical language System (UMLS) \cite{bodenreider2004unified}) to ensure consistency. Language models used tools like TokenMonster\footnote{\href{https://github.com/alasdairforsythe/tokenmonster}{https://github.com/alasdairforsythe/tokenmonster}}, a subword tokenizer and vocabulary generator, and BatchBPE \cite{morgan2024batching}, an efficient Python-based Byte-Pair Encoding (BPE) method that reduces memory use by merging token pairs in batches. In general, the choice of tokenizer significantly affects the embedding space of the model and the training costs \cite{ali2024tokenizer}. Inspired by efficiency of domain-specific vocabulary and tokenization \cite{gu2021domain}, BioMedLM \cite{bolton2024biomedlm} uses a custom BPE tokenizer to prevent the tokenization of healthcare-specific terms \cite{bolton2024biomedlm}. \citet{kim2024mhgpt} employed both, pretrained tokenizer and custom tokenizer to develop SLM model for mental health analysis. All evaluated LLMs exhibited poor performance on medical code querying, frequently producing imprecise or fabricated codes, underscoring the need for further research before clinical deployment \cite{soroush2024large}. There is pressing need for specialized tokenization strategies for ICD-codes query. 

\textbf{Vocabulary Size}: The vocabulary size is the total number of unique tokens (words, subwords, or characters) that the model can recognize and generate. The larger vocabulary enhance both training efficiency and performance of the LLMs \cite{takase2024large}. In SLMs, a dynamic vocabulary curriculum learning improves pretraining efficiency by adjusting token granularity—using longer tokens for predictable content and shorter ones for complex text—leading to better scaling to vocabulary size \cite{yu2025scaling}. \citet{tao2024scaling} introduced three methods to predict compute-optimal vocabulary sizes for SLMs. However, input and output vocabularies may exhibit distinct scaling behaviors in smaller models, underscoring the need for independent scaling strategies to optimize model design \cite{huang2025over}. 

\subsection{Attention mechanisms}
\label{attention}
Attention mechanisms play a central role in enabling language models to selectively focus on relevant information within input sequences. In this section, we describe core attention architectures and efficient attention mechanisms along with their time complexities and discuss recent efficiency-driven adaptations designed to support SLM development. 

\textbf{Core attention architectures}:
Core attention mechanisms in language models can be optimized for healthcare by investigating how clinical concepts and contextual information interact during processing. The attention mechanism has evolved from self-attention to cross-attention, and further to multi-head, entity-aware, and knowledge-guided variants — each contributing to advancements in healthcare applications. For example, \textit{Self-attention} is useful in attending certain medical aspect— symptoms, diagnoses, or medications, capturing dependencies like the relationship between ``chest pain” and ``ECG findings” \cite{vaswani2017attention}. \textit{Cross-attention} enables one source of information (e.g., lab results) to attend to another (e.g., physician notes), supporting rich understanding by integrating information from both structured and unstructured information \cite{lai2024carzero}. \textit{Global attention} highlights key clinical anchors (e.g., “principal diagnosis” or “chief complaint”) that influence interpretation of the entire record  \cite{liu2025attention}. \textit{Multi-head attention} (MHA) models multiple types of clinical relationships in parallel-temporal patterns, medication-disease interactions, or symptom clusters \cite{meng2025transmla}. The \textit{knowledge-guided attention} integrates biomedical ontologies or knowledge graphs (e.g., UMLS) to bias attention toward medically validated relationships, enhancing clinical accuracy and explainability \cite{li2023beginner}. However, a key limitation of core attention mechanisms—particularly self-attention—is their memory inefficiency, as the computational and memory cost scales quadratically with the input length, making it challenging to process long healthcare-related documents and patient histories in EHR.

\textbf{Efficient attention}: To address computational bottlenecks, SLM in healthcare leverage memory-efficient attention adaptations. \textit{Sparse attention} restricts these interactions to the most relevant information— focusing only on recent or abnormal findings—making it ideal for processing lengthy medical notes efficiently \cite{beltagy2020longformer,child2019generating}. \textit{Multi-query attention} (MQA) allows multiple attention heads to share the same keys and values (KV) pairs \cite{xu2023multi}. For example, in a clinical note, different attention heads can be used to extract aspects like demographic information, medication history, and clinical conditions, while sharing the same KV pairs. 
\textit{Grouped-query attention} (GQA) groups query heads into clusters, with each cluster sharing a KV pair \cite{ainslie2023gqa}. For example, when extracting social determinants of health (SDoH), one cluster of heads might focus on housing, another on employment, and another on education. \textit{Multi-Head Latent Attention }uses low-rank compression of KV pairs to reduce the need for a KV cache \cite{lu2024small}. For example, a model extracting quality of life (QoL) measures from patient feedback can use this compression to handle large datasets without storing full-dimensional attention matrices for each concept. QCQA introduces a quality- and capacity-aware grouped query attention mechanism that optimally balances key-value (KV)-cache size and text generation accuracy, outperforming GQA by up to 20\% in accuracy and reducing memory requirements by 40\% without fine-tuning \cite{joshi2024qcqa}.
\textit{Flash Attention} (FA) divides the attention matrix into smaller blocks and recomputes them during backpropagation, saving up to 4 times more memory without losing precision \cite{dao2205fast}. For example, in a clinical question-answering system, instead of storing the entire attention matrix for a patient's full medical report, FA stores only key blocks and recomputes others as needed, reducing memory usage and allowing the system to handle larger datasets like multi-visit histories.

 \textit{Low-Rank Factorization} approximates the full attention matrix by decomposing it into smaller, lower-dimensional matrices, retaining only the essential relationships. In a clinical concept extraction model, this approach increases efficiency by replacing large KV matrices with more compact representations. A series of advancements in low-rank adaptation techniques have emerged to optimize LLM fine-tuning. 
\textit{Memory-Efficient Attention} optimizes processing by calculating attention scores without storing the entire matrix. In a real-time CDSS, this approach allows for dynamic attention score calculation using medical records, enabling efficient analysis of patient symptoms and medical history without storing full context. 
\textit{Sliding Window Attention} captures local dependencies by focusing on a fixed-sized window of neighboring tokens. The clinical note SUM allows the model to focus on specific sections such as the history of the present illness. 
\textit{Random Projection Attention} uses dimensionality reduction to approximate softmax attention, improving scalability. Building on these foundations, recent advancements have given rise to more sophisticated architectures. Notably, Mamba \cite{gu2023mamba} introduces a selective state space model characterized by input-dependent transitions. In predictive models for patient outcomes, it reduces the complexity of processing high-dimensional data, such as medical tests and demographics, making the model more efficient and computationally cost-effective for large datasets and real-time predictions. We analyzed the time complexity of each attention mechanism to shed some light on how healthcare research can integrate attention-driven component based on available resources in Table \ref{tabattention}.

\begin{table}
\scriptsize

\caption{Time complexities of various attention mechanisms in cNLP, from \emph{least} to \emph{most} efficient. 
Here, \(N\) is the sequence length, \(d\) is the embedding dimension, 
\(M\) is the cross-attended length, 
\(w\) is a local window size, 
\(g\) is the number of global tokens, 
and \(r\) is a low-rank dimension. 
``KG overhead” and \(\epsilon\) indicate extra constant factors due to knowledge bases or entity-related modules. These values are based on official reports or estimated 
according to functionality.}
\centering
\begin{tabular}{|l|l|l|}
\hline
\textbf{Attention type} & \textbf{Time complexity} & \textbf{Use case in healthcare}\\
\hline
Self-Attention 
& \(O(N^2 \cdot d)\) 
& Intra-note relationships \\
Cross-Attention 
& \(O(N \cdot M \cdot d)\) 
& Structured + unstructured fusion \\
Global (Sparse) Attention 
& \(O(N \cdot w + N \cdot g)\) 
& Global tokens for long notes \\
Multi-Head Attention 
& \(O(N^2 \cdot d)\) 
& Parallel clinical patterns \\
Knowledge-Guided Attention 
& \(O(N^2) + \text{KG overhead}\) 
& Ontology-driven alignment \\
\hline
Sparse Attention 
& \(\sim O(N \cdot w),\, O(N \log N), \dots\) 
& Restrict to local windows / relevant tokens \\
Multi-Query Attention 
& \(O(N^2 \cdot d)\) [reduced memory] 
& Heads share the same K/V pairs \\
Grouped-Query Attention (GQA) 
& \(O(N^2 \cdot d)\) [fewer K/V sets] 
& Group queries into clusters \\
Flash Attention 
& \(O(N^2 \cdot d)\) [blockwise] 
& Near full precision, reduced memory \\
Multi-Head Latent Attention 
& \(O(N^2 \cdot r), \; (r \ll d)\) 
& Low-rank latent representations \\
Low rank factorization & \(O(N^2 \cdot r), \; (r \ll d)\) & Dimensionality reduction for large clinical corpora\\
Memory efficient attention & \(O(N^2 \cdot d)\) & Handle lengthy EHR data on limited GPU/TPU resources\\
Single window attention & \(O(N \cdot w)\) & Focus on a single local segment (useful for chunk-based analysis)\\
Random Projection Attention
& \(O(N)\)
& Linear approximation of full attention \\
\hline

\end{tabular}
\label{tabattention}
\end{table}

\section{Prompting, structured fine-tuning and reasoning: Adapting SLMs to clinical precision}
\label{icttuning}
After pretraining and the integration of attention mechanisms, prompt engineering, reasoning and instruction tuning becomes the most essential components to adapt the general instruct-language model to specific healthcare tasks. Fine-tuning enables the model to align with domain-specific objectives and improve performance in hospital settings. In this section, we discuss in-context learning (\S\ref{ict}) and fine-tuning (\S\ref{finetuning}).

\subsection{In-context learning}
\label{ict}
\textit{In-context learning} (ICL) in healthcare is the ability of language models to learn and adapt to new healthcare NLP tasks and provide meaningful outputs based on the context in clinical notes, without requiring explicit retraining or fine-tuning.
While \citet{dong2022survey} provide a broad overview of ICL, a comprehensive survey dedicated to its applications in cNLP remains absent. In-context learning enables models to adapt to tasks without gradient updates, while \textit{preference alignment} further refines their behavior to align with human values and domain-specific expectations, ensuring safer and more relevant responses in healthcare.

\textbf{Alignment}: The most renowned method for preference alignment is \textit{reinforcement learning from human feedback (RLHF)}, which relies on reward models and policy updates that often introduce instability due to probability ratio-based optimization \cite{bai2022training}. DeepSeek's \textit{Group Relative Policy Optimization} (GRPO) made RLHF more efficient \cite{shao2024deepseekmath}, especially for complex reasoning tasks, while Stanford’s \textit{direct preference optimization (DPO)} refines clinician response selection by dynamically adjusting the log probabilities of preferred versus non-preferred responses. These models uses human feedback to train language models and generate more preferred responses without requiring a separate reward model. Past studies reported that \textit{supervised fine-tuning} (SFT) (\S\ref{finetuning}) is sufficient for healthcare CLS tasks, whereas direct performance optimization yields superior results for clinical reasoning, and clinical triage \cite{savage2024fine}. Other than DPO \cite{ahn2024note}, the research community has explored \textit{multi-agent approach} of preferential data annotation through multiple rounds of dialogue \cite{zhang2025iimedgpt,gururajan2024aloe} and \textit{reinforcement learning with environmental feedback} \cite{liubest} in healthcare applications. Thus, preferential alignment methods have the potential to alleviate the labor-intensive workload of clinicians. 

\textbf{Prompt engineering}: 
Prompt Engineering is the process of designing and optimizing inputs (prompts) given to the SLMs to generate meaningful, relevant, and actionable outputs in the healthcare domain \cite{liu2023pre}. Recently, the research community reported comprehensive reviews and surveys on the latest advances in prompt engineering for healthcare \cite{wang2025soft}. 
As the choice of prompt can notably influence the accuracy of LLM responses to medical queries \cite{wang2024prompt}, Prompt engineering has emerged as a critical skill for healthcare professionals, enabling more effective interaction with language models \cite{mesko2023prompt}. Building on this, frameworks like MED-Prompt have been developed to construct clinically relevant prompts that guide models in interpreting and extracting information from medical texts \cite{ahmed2024med}. More recently, the introduction of active prompting—an adaptive technique where the model iteratively refines its responses based on prior feedback—has shown promise in enhancing diagnostic reasoning and minimizing erroneous inferences \cite{diao2023active}. A recent survey by \citet{wang2023prompt} reported a range of methods for designing prompts in healthcare applications.

\textbf{Reasoning}: \textit{Reasoning} further increases the capabilities of model to infer relationships between concepts, drawing conclusions, and applying domain knowledge to produce outputs that are logically consistent with medical standards. 
\citet{huang2022towards} discussed different mechanisms for improving and eliciting reasoning in LLMs and benchmarks methods for evaluating reasoning abilities.
To encourage reasoning in LLMs, chain-of-thought (CoT) prompting  \cite{wei2022chain} and ``let us think step-by-step” prompting \cite{kojima2022large} guides them to generate explicit reasoning steps before answering. CoT allows structured inference by breaking down complex medical queries into intermediate reasoning steps in healthcare NLP tasks \cite{wang2024atscot,jiang2025comt}. \citet{wu2025automedprompt} critiques CoT prompting for its limitations across subspecialties and k-shot approaches for introducing noise, proposing AutoMedPrompt, leveraging textual gradients for optimizing system prompts to enhance medical reasoning in foundation LLMs. 

The integration of external knowledge has advanced healthcare informatics by enhancing factual accuracy and reducing hallucinations \cite{li2025merging, wang2024knowledge}. Retrieval-augmented generation (RAG) further strengthens healthcare language models by incorporating clinical knowledge into text generation to deliver contextually relevant responses  \cite{lahiri2024alzheimerrag}. The integration of RAG and CoT demonstrates improved performance on high-quality clinical notes, where early retrieval effectively anchors subsequent reasoning steps in domain-specific evidence while CoT-driven RAG offers advantages in handling lengthy and noisy notes by guiding retrieval through structured reasoning \cite{wu2025integrating,li2024rt}. In MetaICL, meta-learns task-agnostic prompts for generalizing across specialties (e.g., cardiology to nephrology), ``learning” patterns dynamically from the context \cite{min2021metaicl}.
The integration of structured knowledge from a \textit{knowledge graph (KG)} into the model's training or inference process enhance the model's ability to understand context and reason over relationships. KG has the potential to map patient data to structured ontologies (e.g., SNOMED-CT) and infer relationships (e.g., "fatigue + jaundice → hepatitis risk"). Recent studies on neuro-symbolic AI suggest the need of further investigation of \textit{metacognition}\footnote{model's capacity to monitor, assess, and adapt its own reasoning or output generation processes, often by estimating confidence, recognizing uncertainty, or reflecting on errors.} for advancing the field towards more intelligent, reliable, and context-aware AI systems \cite{colelough2025neuro,didolkar2024metacognitive}. Metacognition improves decision-making in healthcare practitioners through cognitive bias identification, facilitation of systematic approaches, holistic approaches, and creative thinking \cite{church2023does}. MD-PIE emulates cognitive and reasoning abilities in medical reasoning and decision-making \cite{esteitieh2025towards}. However, more robust evaluation frameworks that incorporate metacognitive abilities are required as they are essential for developing reliable language model enhanced CDSS \cite{griot2025large}. 

\subsection{Supervised fine-tuning (SFT)}
\label{finetuning}
Fine-tuning on smaller, task-specific datasets enables language models to effectively leverage pre-trained knowledge for specialized domains or tasks \cite{fu2024biomistral,tran2024bioinstruct}. These techniques address challenges like limited computing resources, data quality, availability, and robustness, ensuring efficient adaptation to new tasks without extensive retraining. Instruction tuning is sometimes used interchangeably with instruction fine-tuning; however, the latter often refers to training with chat-specific templates. While the distinction is debated in the literature, we do not focus on this differentiation in this work.

\textbf{Instruction tuning} leverages instructions for flexibility. Instruction datasets are created by integrating annotated data into instruction-output pairs using templates. Alternatively, LLMs can generate outputs from manually or seed-generated instructions. In the human-generated questions with LLM annotations approach, humans create diverse, relevant questions, while LLMs provide initial answers. These answers are then reviewed and refined by humans, resulting in high-quality datasets for instruction tuning. The LLM-generated SDHG for instruction tuning increases training example diversity and reduces manual effort. \citet{nazar2025design} provide a detailed guidance on designing and evaluating instruction datasets for healthcare is available appropriate data construction methods and strategies ensuring data quality \cite{zhang2023instruction,nazar2025design}. BioInstruct \cite{tran2024bioinstruct} and MIMIC-Instr \cite{wu2024instruction} are the high-quality healthcare instruction datasets designed to enhance the training of language models for clinical NLP tasks.

\textbf{Instruction fine-tuning} leverages focused instructions for specific tasks. A recent study suggests that while full parameter instruction fine-tuning can enhance task-specific performance, it may also lead to knowledge degradation and an increased risk of hallucination, potentially due to the overwriting of pre-trained knowledge during the fine-tuning process \cite{ghosh2024closer}. AlpaCare is fine-tuned on MedInstruct-52k, a diverse machine-generated medical instruction fine-tuning dataset, that outperforms previous medical LLMs \cite{zhang2023alpacare}. LlamaCare enhances LLMs' ability to predict outcomes by fine-tuning existing LLM on various clinical tasks \cite{li-etal-2024-llamacare}. Instruction fine-tuning is employed to adapt the SLM for biomedical tasks \cite{luo2024taiyi}, medical entity recognition \cite{liu2024evaluating}, radiology tasks \cite{ranjit2024rad}, ICU readmissions \cite{namvar5098278predicting}, medical documentation \cite{leong2024efficient} and billing and coding \cite{rollman2025practical}. However, the high-quality, diverse, and relevant instructions are key for SLM fine-tuning. 

\textbf{Parameter efficient fine tuning (PEFT)} adapts models to specific tasks by training only a small (upto <5\%) subset of parameters making them deployable on edge devices such as tablets in rural clinics. This rapid customization fine-tunes the model more efficiently while preserving the advantages of the pre-trained model. Starting with LoRA, subsequent methods like LoRA+, LoRA-FA, and AdaLoRA improve efficiency, memory usage, and adaptability, while approaches such as LoRA-drop, and DoRA introduce novel strategies like random matrix adaptation, output-based pruning, and weight decomposition to further enhance parameter-efficient tuning \cite{mao2025survey}.
\textit{LoRA} injects low-rank matrices into existing weights, allowing models to adapt to specialized tasks without full retraining, preserving foundational medical knowledge with minimal overhead \cite{shi2024mentalqlm}. Recent fine-tuning of Phi3.5-mini\footnote{\href{https://huggingface.co/microsoft/Phi-3-mini-4k-instruct}{https://huggingface.co/microsoft/Phi-3-mini-4k-instruct}} through quantized low-rank adapters (QLoRA) underscores its potential for providing basic medical recommendations in telemedicine and chatbot systems.
\textit{Adapter modules}, inserted between transformer layers, facilitate multi-task learning on these small layers instead of modifying the base model as mixture of experts \cite{cai2025survey}. Domain-specific PEFT approach \cite{gajulamandyam2025domain} is employed for hospital discharge paper SUM \cite{goswami2024parameter}, clinical RE \cite{zhou2023llm}, and biomedical NER \cite{abishek2024bio}. \textit{Prompt tuning} optimizes continuous ``soft prompts” to guide models toward structured outputs \cite{wang2023medical,he2024prompt}. In \textit{prefix tuning} a small set of trainable parameters (prefixes) is prepended to the input representation in a frozen language model, allowing adaptation to domain-specific tasks without modifying the core model weights. 

\section{Compression techniques: Ensuring accessibility and sustainability}
\label{compression}
The need for compressed models—via quantization, pruning, and knowledge distillation—arises from the demand to deploy LLMs efficiently on resource-constrained devices, reduce inference latency and memory usage, cut energy costs, and enable real-time applications while maintaining performance close to their larger counterparts. In healthcare, compressed models enable real-time CDSS, preserving patient privacy while operating on limited hardware for NLP-driven clinical decision-making. 

\subsection{Knowledge distillation}
\label{knowdistil}
Knowledge distillation (KD) is a model compression technique that trains a smaller \textit{student} model to replicate the behavior of a larger \textit{teacher} model~\cite{hinton2015distilling}. KD has proven effective for creating high-performance SLMs from LLMs while significantly reducing model size \cite{wang2024comprehensive}. KD is comprised of two components: knowledge and distillation. The knowledge covers a variety of teacher-student frameworks: labeling (teacher-generated outputs), expansion (in-context sample generation), data curation (meta-driven synthesis), feature extraction (teacher-derived internal representations), feedback (evaluative guidance on student outputs), and self-knowledge (student-led output generation and quality filtering) \cite{xu2024survey}. Distillation covers SFT, divergence and similarity, Reinforcement learning, rank optimization. In \textit{SFT}, also known as sequence-level KD, the student model is trained to maximize the likelihood of teacher-generated sequences, aligning its predictions with the teacher’s outputs through divergence or similarity-based loss functions. \textit{Similarity-based methods} aligns internal representations of teacher and student models while \textit{divergence-based methods} minimize distributional differences. \textit{Reinforcement learning-based distillation} leverages teacher-generated feedback to train reward models and optimize student policies, aligning student outputs with teacher preferences through reward maximization and divergence minimization. \textit{Ranking optimization} offers a stable, sample-efficient alternative to reinforcement learning by directly fine-tuning language models to favor preferred outputs over less favored ones using fixed teacher-generated preference datasets. Furthermore, the KD methods are categorized into \textit{white-box} vs. \textit{black-box} based on the information accessible from the teacher~\cite{yang2024survey}. 

In \textbf{white-box KD}, a student not only observes the teacher’s outputs but also has access to the teacher model’s internal parameters or representations~\cite{agarwal2024policy}. This additional insight into the teacher’s ``internal structure” often enables the student to perform better since it can directly leverage the teacher’s learned features and hidden knowledge, addressing the challenges of LLM scalability and portability for health event prediction~\cite{ding2024distilling}. A feature-level knowledge distillation technique was introduced for medication recommendation, thereby training the SLM using hidden states of LLM~\cite{liu2024large}. In \textbf{black-box KD}, by contrast, only the teacher’s outputs (e.g. predicted logits or generated responses) are available to the student; the ``internal structure” of the teacher is treated as a \textit{black box}~\cite{peng2023instruction,wang2023scott}. This scenario commonly arises when using proprietary LLM APIs as teacher models, where only input–output pairs are accessible, but the underlying model weights remain unavailable. Despite the limited information, black-box KD has shown promising results by fine-tuning students on the input–output behaviors of large models.
Complex reasoning and problem-solving skills can be transferred by distilling the teacher’s CoT and prompt-based capabilities. 

\textbf{Discussion on KD}: Both black box and white box KD mechanisms are widely used in healthcare for developing SLM. Some notable studies are employing white-box KD for medication recommendation \cite{liu2024large}, health event prediction, lung cancer detection \cite{pavel2024non}, emotion recognition in daily healthcare \cite{zhang2024llm}, clinical IE \cite{vedula2024distilling} and medical data SUM \cite{liu2024enhancing}. A sleep health management SLM leverages few-shot CoT (black-box) distillation to transfer problem-solving strategies, long-tail expert knowledge, and personalized recommendation capabilities from LLMs into the SLM ~\cite{zheng2024sleepcot}. Key problems identified while deploying knowledge distillation in the medical domain are insufficient training data, lack of data sharing, long-tail data distribution, label imbalance, and potential for further performance improvement \cite{meng2021knowledge}. In Adapt-and-Distill, both the teacher (LLM) and the student (SLM) are first adapted or fine-tuned to the target domain or task distribution, followed by knowledge distillation~\cite{yao2021adapt}. A task-agnostic, domain-adaptive distillation approach creates small, fast, and effective models by expanding domain-specific vocabulary and compressing large pre-trained models.

\subsection{Pruning}
\label{pruning}
Pruning is a long-established technique for model compression that reduces the size of a model by removing unnecessary or less important parameters~\cite{lecun1989optimal,han2015learning}. In language models, many weights can be zeroed out or eliminated with little to no effect on overall performance \cite{bergsmastraight}. By pruning these redundant parameters, we can significantly shrink model size and accelerate inference, with only a minor drop in accuracy if done carefully. Pruning thus makes models more storage-friendly, memory-efficient, and computation-efficient. Based on the granularity of removal, pruning is categorized as \textit{unstructured pruning} and \textit{structured pruning}.

\textbf{Structured pruning} streamlines LLM by eliminating entire architectural components—such as neurons, filter channels, attention heads, or even layers—instead of removing individual parameters~\cite{ma2023llm,xia2023sheared}. By selectively pruning structurally coherent groups of parameters, this approach yields a smaller yet regularly structured model—maintaining dense connectivity while reducing inference costs, for instance, LLM-Pruner~\cite{ma2023llm}.
\textbf{Unstructured pruning} removes individual weights from LLM without considering any specific structure within the model~\cite{frantar2023sparsegpt}. Unstructured pruning removes less important weights based on a threshold, resulting in a sparse model where the architecture remains unchanged, as long as enough weights stay to maintain performance. Unstructured pruning can deliver higher accuracy for a given compression ratio by allowing more targeted weight removal. However, its irregular sparsity pattern leads to inefficient memory access and limited speed-ups on standard hardware. To fully leverage the compression benefits, specialized libraries or hardware are needed, and even then, acceleration gains are modest unless the model is highly sparse. Post unstructured pruning, significant retraining or fine-tuning is often required to recover lost accuracy that can be resource-intensive.

\textbf{Discussion on pruning}: Contextual pruning is a promising method for building domain-specific language models \cite{valicenti2023mini}. Recent research has focused on one-shot pruning methods and improved criteria to minimize the need for costly retraining~\cite{zhang2024pruning}, validating its performance for healthcare tasks. ATP (All-in-One Tuning and Structural Pruning) is a unified approach that dynamically identifies optimal substructures during fine-tuning, integrating structural pruning with LoRA-aware regularization, achieving superior performance recovery compared to two-stage pruning methods in domain-specific applications \cite{lu2024all}. The research community has witnessed a range of pruning mechanisms for generic language models \cite{kim2024efficient}, however, its investigation for healthcare domain remains unexplored. 

\subsection{Quantization}
\label{quantization}
Quantization is the process of reducing precision of the model's weights and activations from floating-point values (e.g., 32-bit) to lower-bit representations (e.g., 8-bit), thereby drastically reducing the memory footprint and computational cost of the model~\cite{gray1998quantization}. Quantization reduces model size by up to 4x (for 8-bit) and accelerates inference by using efficient low-bit operations. While it may cause slight precision loss due to mapping floating-point values to discrete levels, careful techniques can minimize accuracy degradation. Three components for quantization are weight, activation, and fixed-point \cite{bai2024beyond}. Quantization is categorized based on the timing of quantization in the model's lifecycle: \textit{Post-Training Quantization (PTQ)} and \textit{Quantization-Aware Training (QAT)}.

\textit{Post-training Quantization} (PTQ) compresses a model after it has been fully trained by quantizing its weights (and optionally activations) to a lower precision, without requiring any additional retraining~\cite{dettmers2022gpt3}. Post-Training Quantization (PTQ) is a simple and efficient method where an already trained LLM is converted to lower precision (e.g., 8-bit) with minimal data, without requiring full fine-tuning. It is faster and more cost-effective than retraining but may compromise on accuracy due to quantization errors.
In practice, PTQ methods use an engineering approach to reduce accuracy loss. \textit{Quantization-Aware Training} (QAT) simulates low-bit precision during training or fine-tuning process, allowing the model to adapt to learn and adjust for quantization errors~\cite{liu2023llm}. This typically leads to better accuracy than PTQ, especially at very low bit widths. \textit{Mixed precision} training in SLMs use both 32-bit (FP32) and 16-bit (FP16 or BFloat16) floating-point formats to optimize performance. Mixed precision allows researchers to train and fine-tune models more efficiently on limited hardware, making it a vital technique for resource-constrained environments and small-scale deployments.

\textbf{Discussion on quantization}: DFQ-SAM is a data-free PTQ approach that calibrates quantization parameters without using original data, helping protect healthcare data privacy during model compression~\cite{li2024privacy}. Quantized LLM were noticed for virtual mental health assistant \cite{kumar2024mental}, and medical documentation \cite{goyal2024healai}. \citet{bolton2024biomedlm} computes precision with low precision but storage and gradient communication with higher precision. A recent study examines two key methods of quantization: Activation-aware Weight Quantization (AWQ), which selectively preserves important weights to maintain performance, and BitsandBytes (BnB), which applies uniform fixed-precision quantization for constructing BioMistral \cite{labrak2024biomistral}. QLoRA-based quantization enables low-precision LLMs to maintain strong reasoning and classification performance, enabling efficient deployment on edge devices for mental health applications \cite{shi2024mentalqlm}.

\section{SLM in healthcare: Outlook}
\label{outlook}
Healthcare professionals expect accurate answers and proactive guidance in articulating their concerns, correct misunderstandings, and ask clarifying questions—achieved through multi-round conversation training to enhance response credibility. Multi-round training is feasible for SLMs than LLMs due to low computational demands, supporting an efficient implementation of on-device reasoning frameworks (e.g., RAG), enabling real-time personalized medical responses. A unified evaluation platform for healthcare SLMs should incorporate diverse benchmark NLP tasks, a well-defined care continuum (diagnosis, treatment), and comprehensive metrics (confidence, safety and factuality) to enable fair comparisons and advancements in the field. Healthcare-specific SLMs are commonly developed using one of three model sources: (i) generic LLMs, (ii) healthcare-specific LLMs, and (iii) generic SLMs. The generic LLMs are pre-trained generic data, serving as a robust backbone for downstream healthcare applications due to their strong linguistic, commonsense and reasoning capabilities. These models are often optimized and fine-tuned on medical corpora for cNLP tasks. Second, healthcare-specific LLMs are pre-trained on medical texts, explicitly designed to understand clinical jargon, biomedical terminology, and the patient-physician communication. The knowledge is further compressed into smaller architectures to form task-optimized SLMs. Third, generic SLMs are inherently lightweight and serve as a compact and computationally efficient foundation. Their adaption to healthcare domain through tuning, aids hospital systems with strict latency and privacy requirements. 
This section discusses the current state-of-the-art (\S\ref{current}), the role of SLM in healthcare (\S\ref{role}), and assessment models (\S\ref{eval}).

\subsection{Current state-of-the-art}
\label{current}
\begin{figure}[!ht]
    \centering
    \begin{subfigure}[b]{0.98\textwidth}
        \centering
        \includegraphics[width=\textwidth]{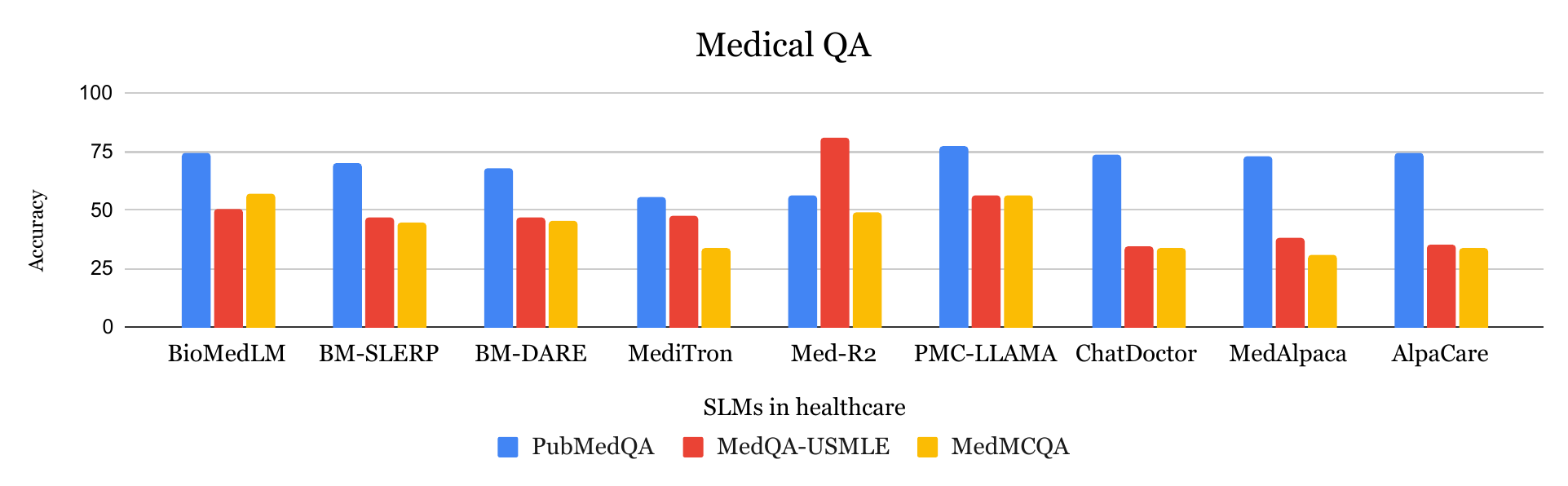}
        \caption{For medical question answering.}
        \label{figmedqa}
    \end{subfigure}
    \hfill
    \begin{subfigure}[b]{0.48\textwidth}
        \centering
        \includegraphics[width=\textwidth]{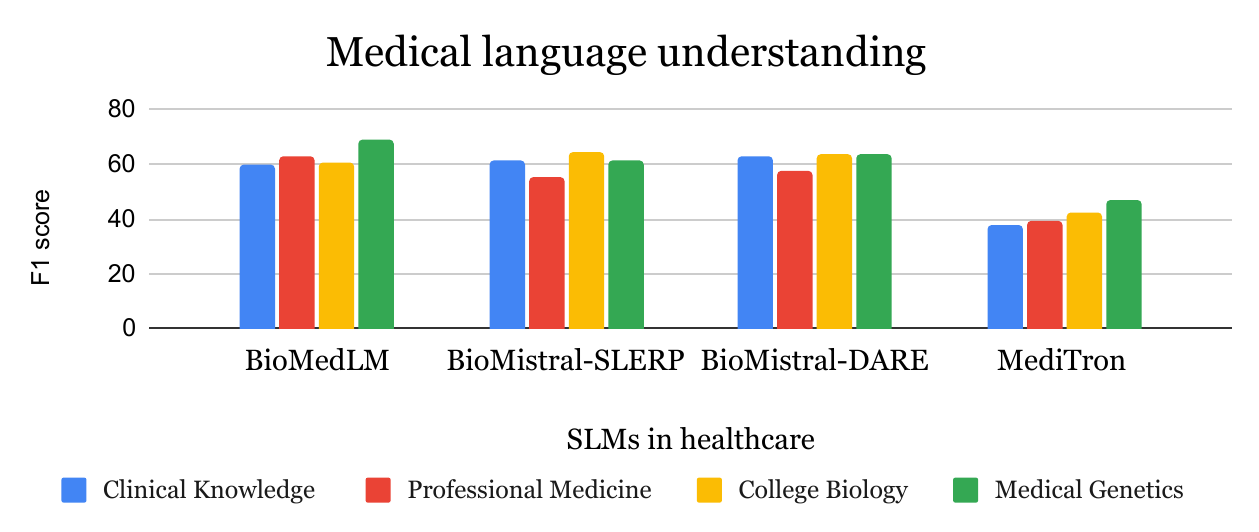}
        \caption{Medical language understanding}
        \label{figlu}
    \end{subfigure}
    \begin{subfigure}[b]{0.48\textwidth}
        \centering
        \includegraphics[width=\textwidth]{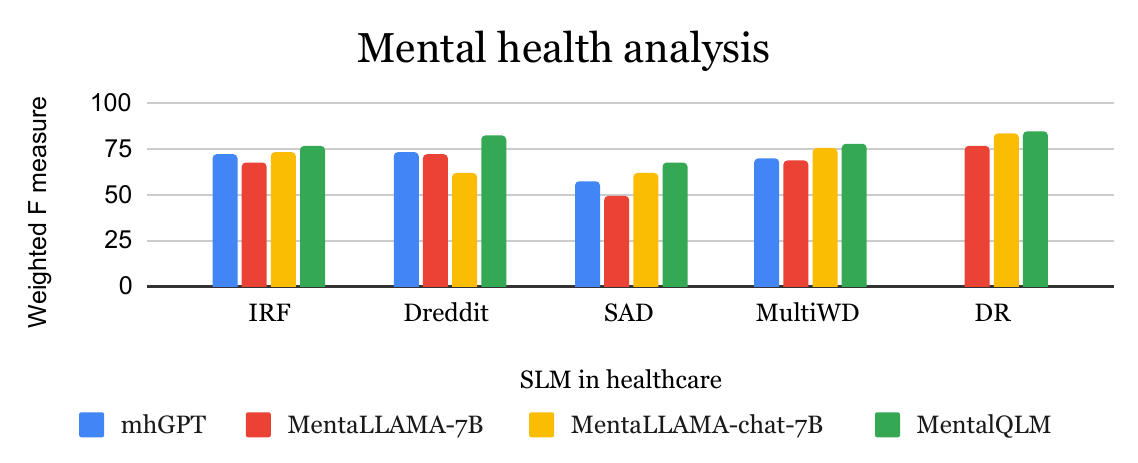}
        \caption{Mental health analysis}
        \label{figmha}
    \end{subfigure}
     \caption{Compiled performance of healthcare SLMs for the most commonly leveraged datasets in healthcare-specific NLP tasks.}
\end{figure}

We assemble the evaluation of existing healthcare SLMs on various NLP tasks, comparing their performance in medical QA through accuracy in Fig.~\ref{figmedqa}, NLU through F1-score in Fig.~\ref{figlu}, and mental health analysis through weighted F-measure in Fig.~\ref{figmha}. We do not claim any computational novelty, but carry over the results from existing studies to report the current state-of-the-art.

\textbf{Medical question answering}: 
Among all others, the results of six SLMs/versions of healthcare-related models developed for medical tasks (BioMedLM \cite{bolton2024biomedlm}, BioMistral-SLERP \cite{labrak2024biomistral}, BioMistral-DARE \cite{labrak2024biomistral}, MediTron \cite{chen2023meditron}, Med-$R^2$ \cite{lu2025med}, PMC-LLAMA \cite{wu2304pmc}) were comparable on three most commonly used medical QA datasets: PubMedQA~\cite{jin2019pubmedqa}, MedMCQA~\cite{pal2022medmcqa}, and MedQA-USMLE \cite{kung2023performance}. In addition to the medical tasks, we collected results from conversational models in healthcare (ChatDoctor \cite{li2023chatdoctor}, MedAlpaca \cite{han2023medalpaca}, and AlpaCare \cite{zhang2023alpacare}). Fig.~\ref{figmedqa} shows no single model dominates across all medical QA datasets. PMC-LLAMA achieves the highest score on PubMedQA (77.09\%), demonstrating strong factual recall in biomedical literature. On MedQA-USMLE, which emphasizes clinical reasoning, Med-$R^2$ significantly outperforms all others with a score of 81.06\%, indicating its robustness in handling complex medical exam-style questions. BioMedLM shows the most consistent performance across all datasets, ranking among the top models for MedMCQA (57.3\%), a multiple-choice QA dataset focused on general medical knowledge. These results underscore the importance of task-specific fine-tuning and diverse pretraining corpora in building high-performing medical language models. The effectiveness of healthcare SLMs varies across datasets, reflecting differences in reasoning complexity and domain specificity. The lack of a universally dominant model highlights the need for adaptive SLM architectures for clinical settings. Incorporating modular training strategies or task-specific adapters could help bridge the performance gap across diverse datasets.

\textbf{Natural language understanding (NLU)}: In Fig.~\ref{figlu}, we collect results for four model (BioMedLM \cite{bolton2024biomedlm}, BioMistral-SLERP \cite{labrak2024biomistral}, BioMistral-DARE \cite{labrak2024biomistral}, MediTron \cite{chen2023meditron},) evaluated on the four common domains of NLU score: clinical knowledge, professional medicine, college biology, and medical genetics \cite{hendrycks2020measuring}. BioMistral-DARE consistently achieves the highest scores across all domains, including Clinical Knowledge (63.1\%), Professional Medicine (57.4\%), College Biology (63.4\%), and Medical Genetics (63.3\%), highlighting its effectiveness in both foundational and applied medical contexts. BioMistral-SLERP closely follows with its strongest result in Medical Genetics (69\%), suggesting robust performance in specialized biological domains. In contrast, MediTron lags across all domains, indicating limited domain adaptation or insufficient training exposure to complex medical content. Fig.~\ref{figlu} highlights the varying effectiveness of models, suggesting the need for SLMs with improved efficiency for hospital utilization.

\textbf{Mental health analysis}: We gather the results from three models: MentaLLAMA \cite{yang2024mentallama}, mhGPT \cite{kim2024mhgpt}, and MentalQLM \cite{shi2024mentalqlm}; among which MentaLLAMA have two variants: \textit{base model} and \textit{chat model}. The more widely adapted datasets are interpersonal risk factors \cite{garg2023annotated}, Dreddit \cite{turcan2019dreaddit}, SAD~\cite{mauriello2021sad}, MultiWD \cite{garg2024multiwd}, and Depression reddit (DR) \cite{pirina2018identifying}. MentalQLM achieves the highest scores on IRF (76.81\%), Dreddit (82.85\%), SAD (67.06\%), and MultiWD (77.88\%), reflecting its robustness in identifying psychological distress, sentiment, and well-being markers as shown in Fig.~\ref{figmha}. For the DR dataset, which lacks a score for mhGPT, MentalQLM narrowly outperforms other models with a top score of 84.20\%. MentaLLAMA-chat-7B also shows competitive performance, particularly in DR (83.95\%) and MultiWD (75.79\%), indicating its strength in conversational contexts. 
The superior performance by existing SLMs underscores the importance of contextual and sensitive language understanding in mental health applications. ChatCounselor is benchmarked against existing LLMs using a robust GPT-4–based evaluation protocol, which emphasizes natural, human-like dialogue over mechanistic response listing \cite{liu2023chatcounselor}. ChatCounselor outperforms most open-source models and approaches ChatGPT in quality, underscoring the impact of domain-specific fine-tuning on enhancing personalized mental health support. Future work should explore benchmarking clinical mental health SLMs against existing mental health–focused or general-purpose LLMs to strengthen the confidence of key stakeholders- healthcare professionals. Moreover, as the training datasets are primarily derived from social media, the development of a clinical model is essential for effective deployment in hospital settings.

\textbf{Miscellaneous}: We further observed three biomedical language models—BioMistral, BioMistral-NLU, and BioGPT—on the \textit{RE task} using the F1 score as the evaluation metric. BioGPT outperforms the other healthcare SLMs. The ClinicalMamba 2.8B achieves an F1 score of 56.51 and 74.34 for \textit{IE task} task of rare and common ICD coding, respectively.
The performance of various language models on \textit{NER tasks} across six biomedical datasets
demonstrates superior performance by BioMistral-NLU. However, the F1 score varies from 55\% to 90\% for different NER dataset, highlight the need of specialized medical NER models. Clinical conditions (e.g., cancer) and medical expertise (e.g., radiology) consider the construction of specialized models for diagnosis generation tasks/RE, and radiology NLI/IE, respectively. The current state of the art highlights the need for language models that achieve an optimal balance between computational efficiency and high predictive accuracy, while ensuring responsible use of resources. 

\begin{table}
\scriptsize
\begin{tabular}{@{}l l l c | c c| c c c |l l@{}}
\toprule
 \textbf{Paper} & \textbf{Year}& \textbf{Base Model} & \textbf{\#Param} & \textbf{ICL} & \textbf{FT} & \textbf{KD} & \textbf{PR} & \textbf{QT} & \textbf{Task} & \textbf{A}  \\
\midrule
BioGPT~\cite{luo2022biogpt} & 2022 & GPT-2 &347M, 1.5B & \ding{51} & \ding{51}  & \ding{55} & \ding{55} & \ding{55} & RE, QA, CLS & \href{https://huggingface.co/docs/transformers/en/model_doc/biogpt}{\ding{51}} \\
BioMedLM \cite{bolton2024biomedlm} & 2024 & GPT-2 & 2.7B & \ding{55} & \ding{51}  & \ding{55} & \ding{55} & \ding{55} & QA, IE, NLU & \href{https://huggingface.co/stanford-crfm/BioMedLM}{\ding{51}} \\
CancerGPT \cite{li2024cancergpt} & 2024 & GPT-2 & 124M & \ding{55} & \ding{51}  & \ding{55} & \ding{55} & \ding{55} & Drug synergy prediction & \ding{55}$^*$ \\
ClinicalMamba \cite{yang2024clinicalmamba} & 2024 & Mamba & 130M & \ding{55} & \ding{51}  & \ding{55} & \ding{55} & \ding{55} & Cohort selection, ICD coding & \href{https://github.com/whaleloops/ClinicalMamba}{\ding{51}} \\
MentalQLM \cite{shi2024mentalqlm} & 2024 & Qwen1.5 & 0.5B & \ding{51} & \ding{51}  & \ding{55} & \ding{55} & \ding{51} & Mental health classification & \ding{55}$^*$ \\
mhGPT \cite{kim2024mhgpt} & 2024 & GPT-NeoX & 1.98B, 2.8B & \ding{55} & \ding{51} & \ding{55} & \ding{55} & \ding{51} & Mental health classification & \href{https://huggingface.co/ChristyBinu-4/mhGPT}{\ding{51}} \\
RadPhi-2 \cite{ranjit2024rad} & 2024 & Phi-2 & 2.7B & \ding{55} & \ding{51}  & \ding{55} & \ding{55} & \ding{55} & Radiology QA, CXR NLP & \href{https://huggingface.co/StanfordAIMI/RadPhi-2}{\ding{51}}  \\
RadPhi-3 \cite{ranjit2024radphi} & 2024 & Phi-3 & 3.8B & \ding{55} & \ding{51}  & \ding{55} & \ding{55} & \ding{55} & summarization, labeling & \ding{55}$^*$ \\

\hline

AlpaCare \cite{zhang2023alpacare} & 2023 & LLAMA-2 & 7B & \ding{51}&\ding{51}&\ding{55}&\ding{55}&\ding{55}& QA, NLU & \href{https://huggingface.co/xz97/AlpaCare-llama2-7b}{\ding{51}} \\
Apollo \cite{wang2024apollo} & 2024 & Qwen & <=7B & \ding{51} & \ding{55} & \ding{55} & \ding{55} & \ding{55} & Multilingual Med QA & \href{https://huggingface.co/FreedomIntelligence/Apollo-7B}{\ding{51}} \\
BioMistral \cite{labrak2024biomistral} & 2024 & Mistral & 7B & \ding{51} & \ding{51} & \ding{55} & \ding{55} & \ding{51} & QA, multilingual med data & \href{https://huggingface.co/BioMistral}{\ding{51}} \\
BioMistral-NLU \cite{fu2024biomistral} & 2024 & Mistral & 7B & \ding{51} & \ding{51} &  \ding{55} & \ding{55} & \ding{55} & NER, RE, DC, QA, NLI & \href{https://github.com/uw-bionlp/BioMistral-NLU}{\ding{51}} \\
CancerLLM \cite{li2024cancerllm} & 2024 & Mistral-7B & 7B & \ding{55}&\ding{51}&\ding{55}&\ding{55}&\ding{55}& Phenotyping and text gen. &  \ding{55}$^*$  \\
ChatCounsellor \cite{liu2023chatcounselor} & 2023 & Vicuna-v1.3 & 7B & \ding{55}& \ding{55}& \ding{55}& \ding{55}& \ding{55}& QA from hospital interviews  & \href{https://github.com/EmoCareAI/ChatPsychiatrist}{\ding{51}}\\
ChatDoctor \cite{li2023chatdoctor} & 2023 & LLaMA-2 & 7B &\ding{55} & \ding{51} & \ding{55} & \ding{55} & \ding{55} & Dialogue, QA, Retrieval & \href{https://github.com/Kent0n-Li/ChatDoctor}{\ding{51}} \\
Med-$R^2$ \cite{lu2025med} & 2025 & Qwen2.5 & 7B & \ding{51} & \ding{51} & \ding{55} & \ding{55} & \ding{55}  & Retrieval + reasoning pipeline & \href{https://github.com/8023looker/Med-RR}{\ding{51}} \\
MedAlpaca \cite{han2023medalpaca} & 2023 & LLAMA & 7B & \ding{55} & \ding{51} & \ding{55}& \ding{55}& \ding{51} & QA and Summarization & \href{https://huggingface.co/medalpaca/medalpaca-7b}{\ding{51}}\\
MentaLlama \cite{yang2024mentallama} & 2024 & LLaMA-2 & 7B & \ding{55} & \ding{51}  &\ding{55} & \ding{55} & \ding{55} & Mental health generation & \href{https://github.com/SteveKGYang/MentalLLaMA}{\ding{51}} \\
Meditron \cite{chen2023meditron} & 2023 & LLaMA-2 & 7B & \ding{55} & \ding{51} & \ding{55} & \ding{55} & \ding{55}  & Med QA, reasoning & \href{https://huggingface.co/epfl-llm/meditron-7b}{\ding{51}} \\
PMC-LLAMA \cite{wu2304pmc} & 2023 & LLaMA-2 & 7B & \ding{51} & \ding{51} &  \ding{55} & \ding{55} & \ding{55} & Medical QA, dialogue & \href{https://huggingface.co/chaoyi-wu/PMC_LLAMA_7B}{\ding{51}} \\

\bottomrule
\end{tabular}

\small \caption{Comparison of existing SLM in healthcare: Here, ICL: In-context learning, FT: Fine tuning, KD: Knowledge distillation, PR: Pruning, QT: Quantization. The tasks are RE: Relation extraction, CLS: Text Classification, QA: Question Answering, IE: Information extraction. A is availability. $^*$ indicates that code is not released as on 1st April 2025.}
\label{tabl:comparison}
\end{table}

\textbf{Discussion on comparative analysis of SLMs}: SLMs have rapidly evolved for a diverse range of healthcare applications. Table \ref{tabl:comparison} summarizes existing healthcare SLMs, detailing their base architectures, adaptation techniques, and target NLP tasks. Most SLMs are built on lightweight architectures like GPT-2, Phi-2, Mamba, Qwen, and LLaMA-2, often adapted with techniques such as fine-tuning and in-context learning.  However, fewer models leverage quantization. Models like BioMistral, BioMistral-NLU, MentalQLM, and PMC-LLAMA exemplify state-of-the-art performance through multi-task adaptability, multilingual benchmarking, and specialized reasoning tasks (e.g., CoT QA). Yet, many efforts are still concentrated on QA and CLS, while areas like clinical SUM, temporal reasoning, and wellness dimension extraction remain underexplored. Additionally, we marked the availability (A) of SLMs in healthcare domain.

At present, healthcare models employ knowledge-aware methods, including ontology grounding using UMLS or SNOMED CT and RAG to enhance factual correctness. Integrating advanced interpretable reasoning mechanisms, such as metacognition, holds promise for enhancing the trustworthiness and reliability of CDSS. There is a pressing need to broaden the scope of tasks beyond traditional disease-centric applications like question answering and classification, toward holistic patient care, including the extraction and interpretation of emotional, social, behavioral, and wellness dimensions. Establishing standardized benchmarks for emerging clinical NLP tasks - wellness assessment, mental health triage, and personalized care planning will be vital to evaluating real-world utility and scientific progress. Moreover, \citet{magnini2025open} evaluated the use of locally deployed open-source generic SLMs for medical chatbots, focusing on hypertension self-management to ensure privacy and reliability. It demonstrates that lightweight SLMs can perform competitively in key tasks, offering a viable, privacy-preserving alternative to large proprietary models.

\subsection{The role of SLM in healthcare}
\label{role}
Beyond the performance of standalone models, we explore the role of SLMs in synergy with LLMs, emphasizing their complementary strengths, capabilities, and applications in healthcare research. The potential of collaborative and competitive roles of SLMs and LLMs is underappreciated in fostering efficient and accessible AI models for CDSS. Amid the broad spectrum of SLM applications observed in prior studies on general models \cite{chen2024role}, this section delineates their specific relevance to NLP-driven decision-making tasks within the scope of healthcare informatics.
\begin{figure}[!ht]
    \centering
    \begin{subfigure}[b]{0.49\textwidth}
        \centering
        \includegraphics[width=\textwidth]{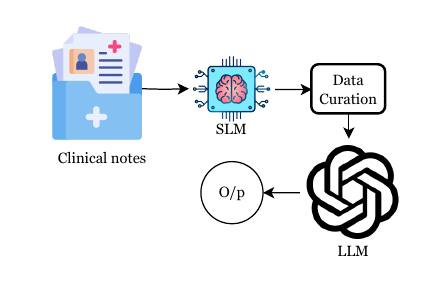}
        \caption{Data curation}
        \label{figrolea}
    \end{subfigure}
    \hfill
    \begin{subfigure}[b]{0.49\textwidth}
        \centering
        \includegraphics[width=\textwidth]{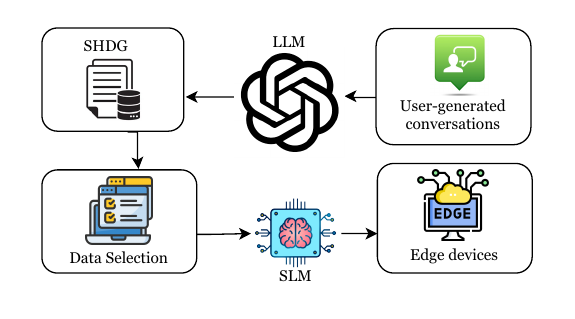}
        \caption{On device personalization}
        \label{figroleb}
    \end{subfigure}
    \begin{subfigure}[b]{0.49\textwidth}
        \centering
        \includegraphics[width=\textwidth]{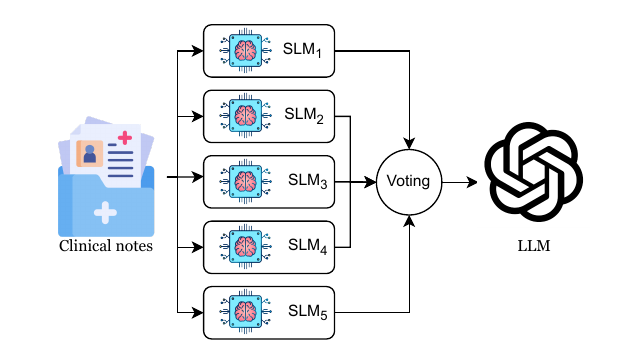}
        \caption{Ensembling}
        \label{figrolec}
    \end{subfigure}
    \hfill
    \begin{subfigure}[b]{0.49\textwidth}
        \centering
        \includegraphics[width=\textwidth]{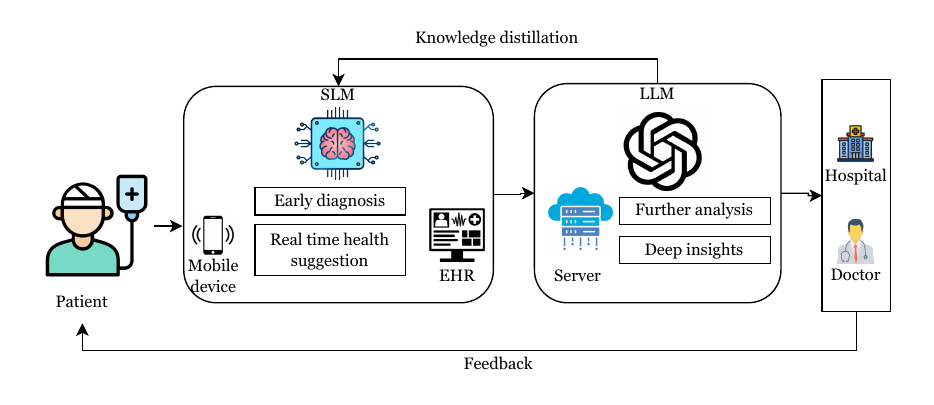}
        \caption{Distillation}
        \label{figroled}
    \end{subfigure}
    \caption{The role of SLM in healthcare}
\end{figure}
\textbf{Data curation}: \citet{zhou2023lima} proved that fine-tuning with just 1,000 high-quality instruction examples can produce a well-aligned model, emphasizing that data quality is more critical than quantity for efficient instruction tuning. Healthcare LLMs are susceptible to hallucinations \cite{longpre2023flan}, highlighting the need of high data quality for evaluation—particularly in terms of factuality, safety, and diversity—prior to model tuning \cite{liumakes}. Leveraging SLMs for efficient and targeted data selection offers a promising strategy to address hallucination \cite{pezoulas2024synthetic,shi2024mentalqlm} (see Fig. \ref{figrolea}). However, despite the use of SLM for SHDG \cite{liu2025generating}, the potential of SLMs in curating quality data remains largely unexplored. 
\textbf{On device personalization.} Patient-generated data often includes sensitive content, making cloud-based annotations undesirable or restricted in clinical settings. Local annotation is possible but must remain sparse to avoid user disruption, and limited device storage hampers large-scale fine-tuning. To this end, \citet{qin2024enabling} proposed the on-device LLM personalization framework that selects and stores representative user data in a self-supervised manner, using minimal memory and occasional annotations (see Fig. \ref{figroleb}). It forms training mini-batches from streaming interactions without retaining the entire data stream. Experiments show superior personalization accuracy and fine-tuning speed over standard baselines.
\textbf{Ensembling SLMs.} \citet{cho2025cosmosfl} introduced a task-level ensemble technique that balances performance and costs by combining SLMs and LLMs, offering Pareto-optimal solutions. The power of ensembling SLMs for healthcare tasks is demonstrated by using multiple fine-tuned SLMs on different clinical report sections to ensure comprehensive coverage \cite{gondara2025elm} (see Fig. \ref{figrolec}). Combined with a LLM for conflict resolution, this approach achieves high accuracy in tumor classification from pathology reports. 
\textbf{Distillation.} Lightweight language models are vital for expanding mental healthcare access, especially in low-resource settings \cite{shi2024mentalqlm}. With the use of knowledge distillation, high-performing cloud-based models can transfer knowledge to smaller edge models, boosting their efficiency and accuracy (see Fig. \ref{figroled}). This enables real-time, personalized support on mobile devices while maintaining low resource demands, addressing global mental health gaps.

\subsection{Assessment methods}
\label{eval}
Challenging the assumption that smaller models inherently compromise accuracy, recent research demonstrates that SLMs have the potential to perform effectively in specialized healthcare tasks \cite{zhang2025rise}.
However, ethical concerns remain critical barriers, underscoring the need for rigorous validation across diverse patient populations to ensure responsible healthcare adoption. As ethical concerns grow, developers and clinicians must proactively mitigate risks to ensure trustworthiness in patient-centered use—especially as SLMs become increasingly personalized and mobile. Evaluating performance is crucial due to the sensitive nature of medical information, where inaccurate or unsafe responses could potentially cause harm.
Healthcare-specific SLM evaluations prioritize domain knowledge correctness through \textit{accuracy and judgment}, \textit{safety} through hallucination control, and alignment with medical standards through \textit{trustworthiness}, and better resource utilization through \textit{memory and energy efficiency}. 

\textbf{Accuracy and Judgment}: \citet{chen2023meditron} developed a novel, physician-curated adversarial question dataset rooted in real-world clinical scenarios, along with a comprehensive 17-metric evaluation framework designed to assess alignment and contextual relevance to clinical practice. Selene Mini is an open-source SLM designed for high-accuracy evaluation across diverse tasks\footnote{\href{https://huggingface.co/blog/AtlaAI/selene-1-mini}{https://huggingface.co/blog/AtlaAI/selene-1-mini}}. Trained with a principled blend of supervised and preference optimization on curated data, it outperforms 11 benchmarks of SLMs as a judge, demonstrating robust, domain-adaptable, and format-resistant performance for real-world deployment. 

\begin{table}
\small
\centering
\caption{Carbon emissions during the pretraining phase of various medical/healthcare LLMs, and the estimated number of trees required to offset one year of those emissions (assuming $0.021\,\text{tCO}_2$ absorbed per tree per year \cite{rainforesttrust2025}). Unmarked emission values are reported by legitimate sources, while values marked with $\approx$ are estimates obtained from base models.}
\label{tab:carbon}
\scriptsize
\begin{tabular}{@{}l l >{\centering\arraybackslash}p{0.2\linewidth}>{\raggedright\arraybackslash}p{0.2\linewidth}@{}}
\toprule
\textbf{Paper} & \textbf{Base Model} & \textbf{Carbon Emissions (Metric Tons CO$_2$)} & \textbf{Eq. \#Trees \textcolor{teal}{\faTree} /Year} \\
\midrule
AlpaCare \cite{zhang2023alpacare} & LLAMA-2 7B & $\approx$0.01 & 1 \\
Apollo \cite{wang2024apollo} & Qwen-7B (v2) & $\approx$30.0 & 1,429 \\
BioGPT~\cite{luo2022biogpt} & GPT-2 Medium (345M) & $\approx$0.10 & 5 \\
BioMedLM~\cite{bolton2024biomedlm} & GPT-2 (2.7B params) & 2.00 \cite{bolton2024biomedlm} & 96 \\
BioMistral~\cite{labrak2024biomistral} & Mistral-7B & 0.074 \cite{labrak2024biomistral} & 4 \\
BioMistral-NLU~\cite{fu2024biomistral} & BioMistral-7B (fine-tuned) & $\approx$0.02 & 1 \\
CancerGPT~\cite{li2024cancergpt} & GPT 2 (124M) & $\approx$0.001 & 1 \\
CancerLLM \cite{li2024cancerllm} & Mistral-7B & $\approx$1.00 & 48 \\
ChatCounsellor \cite{liu2023chatcounselor} & Vicuna-v1.3 7B (fine-tuned) & $\approx$0.01 & 1 \\
ChatDoctor \cite{li2023chatdoctor} & LLaMA-13B (fine-tuned) & $\approx$0.01 & 1 \\
ClinicalMamba~\cite{yang2024clinicalmamba} & Mamba (2.8B) & $\approx$0.04 & 2 \\
Med-$R^2$~\cite{lu2025med} & Qwen 2.5 (7B) & $\approx$0.01 & 1 \\
MedAlpaca \cite{han2023medalpaca} & LLAMA-2 7B & $\approx$0.01 & 1 \\
Meditron~\cite{chen2023meditron} & LLaMA-2 (7B) (continued pretrain) & 0.20\footnote{\href{https://huggingface.co/epfl-llm/meditron-7b}{https://huggingface.co/epfl-llm/meditron-7b}} & 95 \\
MentaLlama~\cite{yang2024mentallama} & LLaMA-2 (7B/13B) & $\approx$0.10 & 5 \\
MentalQLM~\cite{shi2024mentalqlm} & Qwen1.5 & $\approx$0.50 & 24 \\
mhGPT~\cite{kim2024mhgpt} & GPT-based (1.98B) & $\approx$0.10 & 5 \\
PMC-LLAMA~\cite{wu2304pmc} & LLaMA-7B & $\approx$0.08 & 4 \\
RadPhi-2~\cite{ranjit2024rad} & Phi-2 (2.7B) & 0.01 \cite{ranjit2024rad} & 1 \\
RadPhi-3~\cite{ranjit2024radphi} & Phi-3 (3.8B) & $\approx$0.02 & $\sim$1 \\
\bottomrule
\end{tabular}
\end{table}

\textbf{Safety}: A modular safety framework leveraging SLMs was proposed to detect and respond to harmful user queries, mitigating the computational overhead and performance compromises associated with embedding safety mechanisms directly within LLMs \cite{kwon2024slm}. Employing SLMs as standalone safety modules in healthcare offers a scalable, efficient, and language-flexible approach to AI safety \cite{bhimani2025real}. \textit{Safety} infuse medical validation among virtual health assistants to reduce their manipulation leveraged on controlled conversations through question generation, thereby deriving positivity in communications. A pervasive problem with increased reliability on language models is hallucinations which can be classified into factuality and faithfulness. Med-Pal's robust safety alignment showcase its consistent refusal to engage with diverse adversarial prompts, including prompt injection, jailbreaking, misinformation, and harmful outputs—demonstrating effective defense against misuse in clinical dialogue systems \cite{elangovan2024lightweight}. LookAhead tuning enhances the safety of fine-tuned language models by incorporating partial answer previews during training, effectively preserving safety alignment while maintaining strong task performance with minimal computational overhead \cite{liu2025lookahead}. To address the critical gap in evaluating medical safety of LLMs, MedSafetyBench introduces the first benchmark grounded in medical ethics, revealing current models fall short and demonstrating that fine-tuning with this dataset enhances safety without compromising performance \cite{han2024medsafetybench}.

\textbf{Trustworthiness}: Benchmarking LLMs in healthcare requires standardized, domain-specific metrics to ensure safety, accuracy, and ethical integrity while addressing data privacy, bias, and explainability \cite{budler2025brief,aljohani2025comprehensive}. Truthfulness is critical in language models to prevent misinformation. Using the TruthfulQA, BioMistral-7B shows strong performance, outperforming GPT-3.5 Turbo by 4\%, though no model excels across all categories, highlighting task-specific strengths and limitations \cite{labrak2024biomistral}. Faithfulness in AI-generated medical information remains a critical concern, and recent efforts focus on understanding its causes, developing evaluation metrics, and proposing mitigation strategies to ensure factual accuracy across generative medical AI applications \cite{xie2023faithful}.

\textbf{Memory and energy efficiency}: Energy efficiency is a high priority for SLMs, especially when used on battery-powered devices. MELODI is a novel framework and dataset designed to monitor and analyze energy consumption during LLM inference, revealing significant efficiency disparities and offering a foundation for sustainable, energy-conscious AI deployment \cite{husom2024price}. Response length significantly impact power consumption, and concise outputs can reduce energy usage. We measure the reduced environmental impact of healthcare SLM development through carbon footprints in Table \ref{tab:carbon}. Greenhouse equivalences can be calculated with a well-establish calculator\footnote{\href{https://www.epa.gov/energy/greenhouse-gas-equivalencies-calculator}{https://www.epa.gov/energy/greenhouse-gas-equivalencies-calculator}}.

\section{Discussion and conclusion}
\label{discussion}
SLMs are emerging as valuable tools for assisting physicians and patients by enabling personalized treatment plans, accelerating diagnoses, and streamlining medical record analysis. Our investigation of existing NLP-driven SLMs in healthcare—focused on English-speaking countries—aims to support clinical decision-making across the care continuum. We report the search terms and inclusion-exclusion criteria for identifying relevant studies. The timeline of SLM development reveals advances in NLP tasks for specific clinical conditions, healthcare NLP tasks, medical NLP tasks, and doctor-patient conversations/mental health applications. We propose the taxonomies to investigate customization and tuning; the compression techniques for development of specialized SLMs; and outlook for healthcare professionals. We justify the need of SLM development in a high-stake application - healthcare informatics and curate the open-source datasets and models to facilitate future developments. Domain-specific tokenization has advanced healthcare language models by leveraging tools like UMLS, abbreviation disambiguation, and custom BPE methods to preserve medical terminology structure, improving classification accuracy and reducing training costs \cite{hosseini2024leveraging}. The observation - input and output vocabularies exhibit distinct scaling behaviors in small models, raises open questions about optimal vocabulary construction, particularly in healthcare \cite{huang2025over}. How far such optimization can be achieved through data selection and curriculum learning remains an important area for future exploration. Concept-based reasoning in large concept models offers promising potential for future clinical concept extraction by enabling deeper understanding of full clinical statements, including negations and causal relationships \cite{barrault2024large}. We discovered the pressing need to benchmark data selection methods for constructing high quality datasets. 

We discuss architectural foundations for model development and key advancements in SLMs. Given the central role of self-attention in lightweight model architecture, we present a comparative analysis of attention mechanisms based on time complexity to guide the efficient selection and deployment of models for healthcare NLP tasks. Although low-rank adaptation techniques have been explored for SLM development in the past, their application in clinical concept extraction—remains underexplored, presenting opportunities to enhance efficiency and adaptability in task-specific fine-tuning. Advancing trustworthy healthcare SLMs requires integrating human-in-the-loop reasoning, CoT, RAG, knowledge graphs, and emerging meta-cognitive frameworks to improve factual accuracy, contextual understanding, and intelligent decision-making across clinical domains \cite{sethi2025towards,amugongo2024retrieval}. With evolving CoT and RAG frameworks, there is an urgent need to investigate and advance in-context learning and reasoning mechanisms for knowledge-driven healthcare SLMs.  Literature vouch for long training for small models \cite{bolton2024biomedlm}. 

Noticably, many task-specific models were compressed using knowledge distillation, a few healthcare SLM models employ quantization, while pruning remains largely unexplored in this domain. SLMs play a pivotal role in addressing the limitations of LLMs in healthcare by offering efficient, scalable, and privacy-preserving solutions across various stages of clinical AI pipelines. From high-quality data curation and on-device personalization to ensembling and knowledge distillation, SLMs complement LLMs by enabling low-latency, cost-effective, and context-aware decision support. 

Our goal is to provide healthcare professionals with practical insights on deploying SLMs in real-time applications, rather than explaining their internal workings. \textbf{SLMs assist healthcare professionals} by extracting key clinical concepts—such as SDoH, QoL, and therapeutic interventions—from unstructured text. They support CDSS through automated diagnosis assistance, personalized treatment recommendations, and summarization of lengthy patient records. Additionally, SLMs enhance real-time CDSS by identifying rare diseases, drug interactions, and aligning patient data with medical literature. \textbf{SLMs support clinical research} by structuring EHR data and monitoring adverse events, thereby reducing manual chart reviews and enhancing data usability. They automate clinical coding and billing by mapping free-text entries to standardized codes, minimizing administrative burden and claim errors. SLMs also enable clinical QAs by extracting relevant insights from research literature and patient records, accelerating evidence-based investigations. \textbf{SLMs have the potential to enhance patient-centered care} by analyzing sentiment and monitoring mental health to detect risks like worsening depression. They empower virtual health assistants and triage systems, facilitating personalized engagement and real-time conversational support. Additionally, SLMs improve healthcare accessibility through Plain Language Adaptation - simplifying complex medical information for individuals with limited health literacy.

We report our study on the rise of SLMs in healthcare at Github repository. We further investigate and report the reduction in carbon-emission and rise in sustainability due to evolving SLM in healthcare. Further in this section, our concluding remarks covers challenges and future research directions (\S\ref{future}), and the impact of healthcare SLMs (\S\ref{impact}).

\subsection{Challenges and research directions}
\label{future}
The contrast between the theoretical capabilities and real world limitations is a fascinating space to explore. SLMs in NLP-driven healthcare informatics are facing a significant benchmarking challenge. Unlike general-domain NLP, healthcare informatics relies heavily on proprietary, heterogeneous data, limiting reproducibility and transparency. Public datasets offer partial solutions but often lack representativeness. As models grow in complexity and autonomy, evaluation frameworks must evolve to incorporate safety, ethics, and trustworthiness. Standardized benchmarks are essential to ensure consistent assessment, with key criteria including ``doctor-like” communication—demonstrating medical expertise, sound reasoning, and clinically relevant diagnostic insight. Progress requires collaborative efforts, sustainable funding, regulatory alignment, and a workforce skilled in both informatics and health sciences. Edge deployment is critical for expanding access to AI tools in underserved regions, enabling point-of-care decision support, and reducing reliance on cloud-based APIs that may raise data privacy concerns. Interoperability is the model's ability to seamlessly integrate and operate within the existing  Health Information Exchanges (HIEs) from different sites. Another significant advantage of SLMs is their potential when used in \textit{multi-agent AI systems}—collections of specialized agents that collaborate to solve complex tasks. Instead of relying on a single monolithic model like an LLM, multi-agent systems distribute tasks among several domain-specific SLMs. Future research must focus on developing modular interfaces, shared embedding spaces, and universal chaining frameworks that enable plug-and-play interoperability across models of varying sizes and capabilities. We posit the SLM development as a multi-dimensional solution for advancing NLP-centered healthcare informatics. Several active research directions in the development of generic SLMs are being effectively adapted to advance healthcare-specific SLM development. We enlist significant developments in healthcare as follows.

\textbf{Human-in-the-loop systems} integrate SLMs, LLMs, and expert oversight to ensure safe and effective deployment. SLMs assist with triage and uncertainty flagging, while LLMs manage complex synthesis. Embedding human feedback enhances trust and supports continual model refinement. Advancing flexible architectures, evaluation frameworks, and regulatory standards is essential, with future research focusing on efficient evaluators for clinical tasks, emphasizing factuality, safety, and uncertainty \cite{han2024medsafetybench,xie2023faithful,budler2025brief}.
\textbf{Reducing manual chart review.} As LLMs are preferentially aligned with human-preferred responses and surpass human capabilities, the feasibility and reliability of human evaluation diminish, introducing a weak-to-strong generalization,  wherein smaller, less capable models supervise and guide the alignment of more powerful models \cite{burns2023weak}. This paradigm has potential to reduce manual chart review efforts, enabling scalable, cost-effective, safe and factual preference alignment that exceed clinician expertise.
\textbf{Summarizing medical reports.} SLMs perform high-quality abstractive summarization by distilling structured rationales from LLMs \cite{wang2025distilling}. These rationales encapsulate important structural and semantic cues from the document, broken into high-level concepts, text-spans that support those themes, and relationships among key entities in the document. These structured rationales serve as interpretable reasoning traces that guide summary generation, similar to CoT prompting but in a structured, modular format. SLMs are rationale-aware, lightweight summarizers that are trained using intermediate reasoning signals setting a new paradigm for transparent and cost-effective summarization.
\textbf{Bias detection and de-biasing models.} Because SLMs are often fine-tuned on relatively small clinical corpora (e.g., institutional EHRs or specialty-specific datasets), they are at high risk of inheriting data biases or overfitting to local linguistic patterns \cite{sapkota2025comprehensive}. This can manifest in poor generalization to diverse patient populations, inaccurate risk stratification, or propagation of health disparities in AI-driven decision-making. Future research must incorporate domain-aware debiasing techniques such as adversarial training, fairness-aware objective functions, and counterfactual data augmentation. Rigorous bias audits and alignment assessments should become standard in the development lifecycle of healthcare SLMs.
\textbf{Automated Machine Learning (AutoML).} While existing model compression techniques have shown significant progress, they continue to rely heavily on manual design—particularly in defining suitable student architectures for knowledge distillation, which demands considerable human effort. The application of AutoML to LLM remains limited due to the significant computational burden involved in scaling down models with billions of parameters. SLMs address this challenge by evaluating architectural components like model depth and attention head width within a reduced parameter space \cite{liu2024mobilellm}. \citet{shen2024search} reduced the search complexity by incorporating optimized initialization strategies, accelerating convergence. Therefore, the models must integrate AutoML approaches—meta-learning and neural architecture search, that offer a promising direction to mitigate this dependency. This will enhance compression efficiency without compromising performance.

\subsection{Impact of SLMs} 
\label{impact}
SLMs offer infrastructure-friendly deployment on edge devices using optimized hardware such as NVIDIA Jetson or Qualcomm AI accelerators. Software optimization techniques (distillation, quantization and pruning) significantly lower computational demands, enabling model adoption without costly hardware investments. However, challenges such as frequent model updates and deployment across heterogeneous devices may introduce new scalability and maintenance complexities. Given these considerations, investigating the impact of model development is necessary to support responsible application. \textbf{Environmental and economic trade-offs}: Training GPT-3 with 175 billion parameters consumed 1287 MWh of electricity, and resulted in carbon emissions of 502 metric tons of carbon, equivalent to driving 112 gasoline powered cars for a year \cite{patterson2021carbon}. The environmental impact of AI is an escalating concern \citep{rillig2023risks} where SLMs offer a partial solution by reducing the training and inference energy consumption by up to 80\% \cite{liu2024green}. This reduction directly translates into a smaller carbon footprint, making SLMs a more sustainable choice as evident from Table \ref{tab:carbon}. This efficiency is valuable in healthcare, aligning with the growing movement toward green AI, where the need of real-time solutions must be balanced with sustainability goals. The future directions for environmentally conscious green (small/large) language model development in healthcare promotes compression techniques discussed in \S\ref{compression}. \textbf{Ethical and Social Impact.} The integration of SLMs into clinical workflows alleviates provider burden while promoting equitable access, reducing digital disparities, and enhancing patient-centered care. Their lightweight nature enables deployment in low-resource settings, including rural clinics, mobile health units, and edge devices, thereby extending AI-driven support to underserved populations. By lowering the technical and financial barriers associated with large models, SLMs democratize the use of advanced NLP tools across diverse healthcare institutions \cite{mittal2024democratizing}. Additionally, SLMs support data de-identification by enabling efficient, on-device removal of PHI from clinical texts, facilitating privacy-preserving data sharing and compliance with healthcare regulations. SLMs can support patient friendly adoption through plain language adaptation, and empower patients with timely, comprehensible information. However, addressing fairness, bias, and equitable access remains critical \citep{sapkota2025comprehensive}. While the narrower focus of model training datasets allows for more effective bias filtering, it also risks excluding underrepresented groups or contexts. Therefore, careful dataset design is essential to ensure inclusive and responsible AI deployment.
\textbf{Synthesis and Perspectives.}
The goal of SLMs is to prioritize efficiency and accessibility. While, they mitigate environmental, economic, and ethical challenges associated with LLMs, the success of these methods largely depend on interdisciplinary collaborations, for example, hardware engineers optimizing for energy efficiency, clinicians validating medical tools, and policymakers ensuring equitable access. Future work in this direction requires standardized benchmarks for fairness and environmental impact, a step toward sustainable, human-centric AI.

\subsection{Concluding remarks}
This work underscores the transformative potential of SLMs in democratizing AI for healthcare, advocating for models that harmonize performance and ethical responsibility with limited resources. Future efforts must prioritize clinician-AI collaboration and regulatory alignment to unlock scalable, human-centric solutions for NLP-centered clinical applications in English speaking territories. Although LLMs possess extensive medical knowledge, smaller, specialized clinical LLMs outperform them, even when fine-tuned on limited annotated data \cite{zhou2023lima}. However, generic SLM have potential to achieve comparable performance to LLMs in targeted use cases while reducing computational costs, emphasizing the sustainable model development \cite{allal2025smollm2}. Whether task-specific SLM in healthcare achieve comparable performance to LLMs or not, is still an open research question \cite{labrak2024biomistral,fu2024biomistral,zhang2025rise}. Efforts have been made to explore medical QA, language understanding and mental health analysis; a huge gap lies in model development for clinical concept extraction and report summarization. Current research on SLMs remains nascent, with limited exploration of longitudinal real-world deployments. 

While this study focuses on NLP-centered healthcare informatics for English speaking territories, we acknowledge the growing importance of multilingual and non-English models, which is beyond our current scope. Our review excludes multimodal research and the SLMs lacking peer-reviewed evidence. Moving forward, we aim to maintain an up-to-date repository on GitHub to reflect ongoing developments in this evolving field.

\section*{Acknowledgements}
This study was supported by NIH (National Institutes of Health) R01 AG068007.

\bibliographystyle{ACM-Reference-Format}
\bibliography{output}

\end{document}